\documentclass[twoside]{article}

\usepackage[accepted]{aistats2025}
\usepackage[round]{natbib}

\usepackage{amsfonts}
\usepackage{graphicx}
\usepackage{xcolor}
\usepackage{multirow}
\usepackage{caption}
\usepackage{subcaption}
\usepackage{mathtools}
\usepackage{array}

\usepackage{tikz}
\usetikzlibrary{bayesnet}
\usetikzlibrary{shapes.geometric}

\tikzset{latent_var/.style={
  circle,
  draw=black,
  fill=white,
  minimum size=1cm,
  inner sep=0pt,
  text width=1cm,
  align=center,
  scale=0.7
  }
}

\tikzset{latent_var_square/.style={
  diamond,
  draw=black,
  fill=white,
  minimum size=1cm,
  inner sep=0pt,
  text width=1cm,
  align=center,
  scale=0.6
  }
}

\tikzset{small_latent_var/.style={
  circle,
  draw=black,
  fill=white,
  minimum size=0.7cm,
  inner sep=0pt,
  text width=0.7cm,
  align=center,
  scale=0.7
  }
}

\tikzset{est_var/.style={
  rectangle,
  draw=black,
  fill=white,
  minimum size=0.7cm,
  inner sep=0pt,
  text width=0.7cm,
  align=center,
  scale=0.6
  }
}

\tikzset{observed_var/.style={
  circle,
  draw=black,
  fill=black!20,
  minimum size=1cm,
  inner sep=0pt,
  text width=1cm,
  align=center,
  scale=0.7
  }
}

\tikzset{population_var/.style={
  circle,
  draw=black,
  fill=white,
  minimum size=0.8cm,
  inner sep=0pt,
  text width=0.8cm,
  align=center,
  scale=0.5
  }
}

\newcommand{\E}{\mathbb{E}}
\newcommand{\I}{\mathbb{I}}

\allowdisplaybreaks

\begin{document}

%

%

\twocolumn[

\aistatstitle{Factor Analysis with Correlated Topic Model for Multi-Modal Data}

\aistatsauthor{Ma\l gorzata \L az\c{e}cka \And Ewa Szczurek}

\aistatsaddress{Faculty of Mathematics, Informatics \\ and Mechanics, \\University of Warsaw \vspace*{0.1cm}\\Institute of Computer Science,\\ Polish Academy of Sciences \vspace*{0.1cm}\\
\textit{m.lazecka@uw.edu.pl}\And 
Institute of AI for Health, \\
Helmholtz Center Munich \vspace*{0.1cm}\\ Faculty of Mathematics, Informatics \\ and Mechanics, \\University of Warsaw \vspace*{0.1cm}\\
\textit{em.szczurek@uw.edu.pl}}]

\begin{abstract}

Integrating various data modalities brings valuable insights into underlying phenomena. Multimodal factor analysis (FA) uncovers shared axes of variation underlying different simple data modalities, where each sample is  represented by a vector of features. 
However, FA is not suited for structured data modalities, such as text or single cell sequencing data, where multiple data points are measured per each sample and exhibit a clustering structure. 
To overcome this challenge, we introduce FACTM, a novel, multi-view and multi-structure Bayesian model that combines FA with correlated topic modeling and is optimized using variational inference. 
Additionally, we introduce a method for rotating latent factors to enhance interpretability with respect to binary features.
On text and video benchmarks as well as real-world music and COVID-19 datasets, we demonstrate that FACTM outperforms other methods in identifying clusters in structured data, and integrating them with simple modalities via the inference of shared, interpretable factors. 
\end{abstract}
\section{INTRODUCTION}  
Real-world data often spans multiple modalities, each capturing distinct yet complementary aspects of the underlying phenomena. Analysis of such multi-modal data has emerged as a critical machine learning task across various application domains.
 For example, video data combine temporal image sequences, audio signals, and  transcriptions of spoken content, while medical datasets integrate diverse patient measurements ranging from electronic health records to imaging (e.g., computed tomography, CT), and molecular data (e.g., single-cell RNA sequencing, scRNA-seq). Effective synthesis of these heterogeneous information sources is crucial for comprehensive sample characterization and predictive modeling. 

A state of the art approach to interpretable data integration is Factor Analysis (FA) and its multi-view extensions, where each view corresponds to a modality. These methods perform unsupervised factorization of high-dimensional observations into interpretable latent factors and weights.
FA serves multiple analytical objectives: it reduces data dimensionality, enables integration of heterogeneous data sources, and reveals principal axes of variation across samples, facilitating interpretation of complex datasets.

Despite these advantages, FA-based approaches face several fundamental limitations. First, FA models are restricted to handling {\textit{simple}} data, where for each data view the samples are described by vectors. However,  frequently, apart from simple views, multi-modal data views fall into a category of {\textit{structured}} views, where each sample consists of a set of data points. These data points arise from an underlying cluster structure and are characterized by values 
assigned to observed objects. For such structured views, each sample can be summarized by a vector of cluster abundances. Such structured data arise in  various applications, including text document analysis, where documents contain multiple sentences composed of words. Sentences cluster into topics, and documents are represented by topic distributions. Another example is scRNA-seq data, where samples contain multiple cells with gene expression values. Here, cells cluster into distinct cell types, and samples are characterized by cell-type proportions. 
Despite the abundance of such real-world data, there exists no FA-based approach capable of unified modeling of simple and structured views. 
The second limitation of FA is lack of full identifiability in terms of latent factors, as they may be permuted, their signs may be switched, or the weight and factor matrices may be rotated without affecting the model likelihood. 

To address these challenges, we introduce FACTM, a novel Bayesian method designed for joint modeling of both simple and structured data across multiple views. FACTM extends the multimodal FA by leveraging the Correlated Topic Model (CTM), originally developed for text mining, to identify clusters and their prevalences within structured views. 
To enable integration across structured and simple views, we link the FA and CTM parts of the model through dedicated variables that are interpreted as sample-specific modifications to the population-level cluster proportions. 
To address the rotation invariance issue in a way that enhances model identifiability, we propose a supervised rotation method that incorporates additional features with which the  factors are expected to be associated. 


Our contributions are as follows: (i) we propose a novel model capable of handling both simple and structured views; (ii) for structured views, our model infers a covariance matrix that reveals relationships between the identified clusters; (iii) we introduce a method for meaningful supervised rotation, that enhances interpretability of latent factors in FA models; and (iv) through simulations, benchmark datasets, and real-world data, we demonstrate that FACTM outperforms existing methods and we show its practical utility.

\section{RELATED WORK}

\paragraph{Factor analysis}

FA~\citep{fa} is a statistical method of representing data through latent factors, with probabilistic PCA~\citep{ppca} constituting a well-known example. 
An extension of FA suitable for high-dimensional data is the Tucker decomposition, which factors a tensor into component tensors. However, the Tucker decomposition is constrained by the requirement that the number of features in each view must be equal, rendering it inapplicable when these dimensions differ.

Numerous generalizations of FA have been developed to handle multiple data modalities, provided as separate data views with differing numbers of features, among which Bayesian approaches have proven particularly successful. Notable examples include
 Group Factor Analysis (GFA)~\citep{gfa} and Multi-Omics Factor Analysis (MOFA)~\citep{mofa}. Both models use automatic relevance determination to enforce factor-wise sparsity (allowing some factors to be inactive across some views). Additionally, MOFA employs spike-and-slab prior to shrink individual loadings to zero. Recent advancements, such as BASS~\citep{bass} and MuVI~\citep{muvi}, introduce structured sparsity assumptions or  domain-informed priors for the weights, while MEFISTO~\citep{mefisto} extends MOFA to account for spatio-temporal dependencies. The addition of priors limits FA non-identifiability as they impose constraints on the possible weight and factor matrices. 

We emphasize that the introduction of sparsity-inducing priors is crucial as it limits FA non-identifiability by imposing constraints on the possible weight and factor matrices. Additionally, these priors serve as an effective tool for denoising the data. Apart from FA, priors enforcing sparsity were shown to enhance performance in deep neural networks and were successfully applied in recent Variational Autoencoder models (e.g. \cite{tonolini, fallah})

The optimization strategies of the Bayesian FA models are based on  Gibbs sampling or variational inference, which ranges from analytically derived EM-like updates of the variational parameters to automated variational inference methods.

\paragraph{Topic models}
Topic models are widely used methods of unsupervised learning of text documents representations, based on a clustering of words into topics. 
Latent Dirichlet Allocation (LDA)~\citep{lda}, a standard topic model, assumes that the words in each document are drawn from a mixture of topics, which are shared across all documents and defined as distributions over words. The topic proportions are document-specific and are generated from a Dirichlet distribution. We highlight two extensions of LDA. The first, Correlation Topic Model (CTM)~\citep{ctm}, uses more flexible distribution for the topic proportions to enable the inference of a covariance structure among the topics, allowing for the presence of one topic to be correlated with the presence of another. The second extension, ProdLDA~\citep{prodlda}, replaces the LDA's mixture of topics with a weighted product of experts.
Apart from text, topic models were applied also to molecular biological datasets, for example to identify tissue microenvironments~\citep{spatialLDA} or deconvolve cell types from multi-cellular pixel resolution data in spatial transcriptomics~\citep{stdeconvolve}.

\section{DESCRIPTION OF METHODS}


\subsection{Background}
\label{section_background}
\paragraph{Factor analysis} A standard FA model aims to linearly reduce dimensionality of data while preserving the main axes of variation by factorizing a single given data matrix 
$Y \in \mathbb{R}^{N\times D}$, into two matrices: 
$Z \in \mathbb{R}^{N\times K}$ with latent factors, and 
$W \in \mathbb{R}^{D\times K}$ with weights (factor loadings), where $N$ denotes the number of samples, $D$ the number of features, and $K$ the number of latent factors. This relationship can be expressed as $Y = ZW' +\varepsilon$, where $\varepsilon$ captures random noise. We consider the data modality $Y$ in standard FA as   {\textit{simple}}, since each sample $n$ is described by a vector of $D$ features, $Y_n \in \mathbb{R}^D$.

\paragraph{FA invariance}
Let $R\in \mathbb{R}^{K\times K}$ be a rotation matrix satisfying the property $R^{-1}=R'$. The likelihood in factor analysis is invariant under such rotations. After applying a rotation to both latent factors $\tilde{Z} = ZR$ and loadings $\tilde{W} = WR$, the likelihood remains unchanged, as
\[ZW' = ZR R'W' = \tilde{Z}\tilde{W}'.\]
The marginal distributions of $Z$ and $W$ are also invariant under isotropic normal priors. Moreover, even without  assuming an isotropic prior on the loading matrix $W$, there is no guarantee that the variances are indeed unequal. 
Common approaches to address this issue are introducing sparsity constraints (e.g. \cite{mofa, muvi}) or applying specific rotations, such as the varimax rotation \citep{varimax}.

 We note that the ordering and sign of latent factors is also non-identifiable. However, a change in sign does not affect the interpretation of the FA model, and the factors are usually  sorted after the model is fitted by the total variance that they explain across all views.
 
 \paragraph{Variational inference} 
 Variational inference proceeds by maximizing an evidence lower bound (ELBO) of the marginal log-likelihood $p(Y)$
\begin{multline*}
    ELBO(q) := \mathbb{E}_q\log p(X,Y) - \mathbb{E}_q\log q(X)\\
    =\log p(Y) - D_{KL}(q(X)|| p(X|Y)) \leq \log p(Y),
\end{multline*}
 over a family of variational distributions $q(X)$ approximating $p(X|Y)$. Here, $X$ and $Y$ represent hidden and observed variables, respectively, and $D_{KL}$ denotes Kullback–Leibler divergence. Under mean-field assumption 
 the optimal variational distribution $\hat q_i$ that maximises the ELBO can be calculated as follows
\begin{equation}
\label{eq_update}
    \log \hat q(x_i) \propto \mathbb{E}_{-x_i} \log p(X,Y),
\end{equation}
where  $\mathbb{E}_{-x_i}$ denotes the expected value with respect to the $q$ distribution for all the variables $X$ except for $x_i$. The standard approach is to use coordinate ascent, iteratively  updating each variational parameter one at a time until ELBO convergence. For details see \cite{vi_intro} and \cite{vi_book}.

 \subsection{FACTM}
 \label{section_model_description}
 \begin{figure*}
 \centering
\begin{tikzpicture}
\node[observed_var](y){$\overline{y}_{n,i,j}$};
\node[latent_var,  left=1.5 of y, yshift=0cm]  (xi)   {$\xi_{n,i}$}; %
\node[latent_var,  left=1.5 of y, yshift=1.7cm] (beta)   {$\beta_{l}$}; %
\node[latent_var,  left=1.5 of xi, yshift=0cm]  (eta)   {$\eta_{n}$}; %
\node[latent_var,  left=1.5 of eta, yshift=0cm]  (mu)   {$\mu_{n,l}$}; %
\node[latent_var,  left=1.5 of mu, yshift=1.7cm]  (w_l)   {$\overline{w}_{l,k}$}; %
\node[latent_var,  left=1.5 of mu, yshift=-1.5cm]  (z)   {$z_{n,k}$}; %
\node[observed_var,  left=1.5 of eta, yshift=-3cm](y_m){$y_{n,d}^m$};
\node[est_var,  left=0.5 of eta, yshift=0.7cm](mu0){$\mu^{(0)}$};
\node[est_var,  left=0.5 of eta, yshift=1.7cm](sigma0){$\Sigma^{(0)}$};
\node[latent_var_square,  left=1.5 of y_m, yshift=-1.5cm]  (w_m)   {$w_{d,k}^m$}; %
\node[small_latent_var,  left=0.5 of y_m, yshift=0cm]  (tau_m)   {$\tau_{d}^m$}; %
\node[small_latent_var,  left=of w_m, yshift=-0.6cm]  (w_tilde)   {$\tilde{w}_{d,k}^m$}; %
\node[small_latent_var,  left=of w_m, yshift=0.6cm]  (s)   {$s_{d,k}^m$};
\node[small_latent_var,  left=of s, yshift=0cm]  (theta_m)   {$\theta_{k}^m$}; %
\node[small_latent_var,  left=of w_tilde, yshift=0cm]  (alpha_m)   {$\alpha_{k}^m$}; %
\node[small_latent_var,  left=of w_l, yshift=0cm]  (alpha)   {$\overline{\alpha}_{k}$}; %

  \node[const, left=0.65 of beta, yshift=0cm] (par_beta) {$\alpha_0^{\beta}$};
 \node[const, left=0.65 of alpha_m, yshift=0.2cm] (par_alpha_m_a) {$a_0^\alpha$};
   \node[const, left=0.65 of alpha_m, yshift=-0.3cm] (par_alpha_m_b) {$b_0^\alpha$};
   \node[const, left=0.65 of theta_m, yshift=0.3cm] (par_theta_m_a) {$a_0^\theta$};
   \node[const, left=0.65 of theta_m, yshift=-0.2cm] (par_theta_m_b) {$b_0^\theta$};
   \node[const, left=0.65 of alpha, yshift=0.3cm] (par_alpha_a) {$\overline{a}_0^\alpha$};
   \node[const, left=0.65 of alpha, yshift=-0.3cm] (par_alpha_b) {$\overline{b}_0^\alpha$};
   \node[const, left=0.65 of mu, yshift=0cm] (par_t) {$t$};

  \edge[->] {xi} {y} ;
  \edge[->] {beta} {y} ;
  \edge[->] {par_beta} {beta} ;
  \edge[->] {eta} {xi} ;
  \edge[->] {mu} {eta} ;
  \edge[->] {mu0} {eta} ;
  \edge[->] {sigma0} {eta} ;
  \edge[->] {w_l} {mu} ;
  \edge[->] {z} {mu} ;
  \edge[->] {par_t} {mu} ;
  \edge[->] {z} {y_m} ;
  \edge[->] {w_m} {y_m} ;
  \edge[->] {w_tilde} {w_m} ;
  \edge[->] {s} {w_m} ;
  \edge[->] {tau_m} {y_m} ;
  \edge[->] {theta_m} {s} ;
  \edge[->] {alpha_m} {w_tilde} ;
  \edge[->] {alpha} {w_l} ;
  \edge[->] {par_alpha_m_a} {alpha_m} ;
  \edge[->] {par_alpha_m_b} {alpha_m} ;
  \edge[->] {par_theta_m_a} {theta_m} ;
  \edge[->] {par_theta_m_b} {theta_m} ;
  \edge[->] {par_alpha_a} {alpha} ;
  \edge[->] {par_alpha_b} {alpha} ;
  
  \plate[yscale=1, yshift=0cm] {dplate} { %
    (y_m)(w_m)(tau_m)
  } {\scriptsize{$d=1,2,\ldots, D^m$}} ;
  \plate[yscale=1, xscale=1.05] {mplate} { %
    (y_m)(w_m)(tau_m)(theta_m)(alpha_m)(dplate.north west)
  } {} ;
  \node[anchor=north west] at (mplate.north west) {\scriptsize{$m=1,2,\ldots, M$}};
  \plate {jplate} { %
    (y)
  } {\scriptsize{$j=1,2,\ldots, J_i$}} ;
  \plate {iplate} { %
    (y)(xi)(jplate.north west)(jplate.south east)
  } {\scriptsize{$i=1,2,\ldots, I_n$}} ;
  \plate[yscale=1.05, yshift=0.2cm] {lplate} { %
    (beta)(mu)(w_l)
  } {} ;
    \node[anchor=north east] at (lplate.north east) {\scriptsize{$l=1,2,\ldots, L$}};
    \plate[yscale=1, yshift=0cm] {kplate} { %
    (theta_m)(alpha_m)(z)(alpha)(mplate.north west)(par_alpha_b)(lplate.north west)
  } {} ;
  \node[anchor=north west] at (kplate.north west) {\scriptsize{$k=1,2,\ldots,K$}};
  \plate[yscale=0.93, yshift=4] {nplate} { %
    (z)(y_m)(mu)(eta)(xi)(y)
    (iplate.north east)(lplate.south west)(dplate.south west)
  } {} ;
  \node[anchor=south east] at (nplate.south east) {\scriptsize{$n=1,2,\ldots, N$}};

  \plate[yscale=1, yshift=0, color=red, style=dashed] {FA} { %
    (mu)(par_alpha_m_b)(mplate.north east)(kplate.north west)
  } {} ;
  \node[left=0.3of z, yshift=0.5cm]  {\textcolor{red}{\parbox{3cm}{Multimodal \\ Factor Analysis}}};
  \plate[yscale=0.95, yshift=2, color=blue, style=dashed] {CTM} { %
    (mu)(y)(beta)(lplate.north east)(iplate.south east)
  } {} ;
  \node[anchor=north east] at (CTM.north east) {\textcolor{blue}{\parbox{2cm}{Correlated \\ Topic Model}}};
\end{tikzpicture}
\caption{Graphical representation of FACTM.
A single structured view is shown (in blue), although any number is possible.}
\label{fig_factm_plate}
\end{figure*}
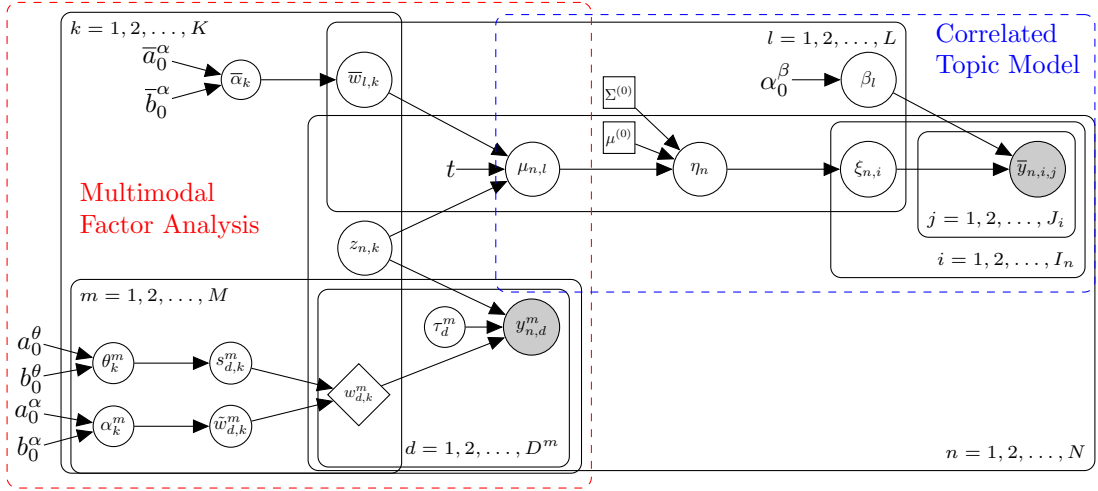
FACTM (Fig.~\ref{fig_factm_plate}) extends FA for multiple simple and structured views. For simple data modalities, FACTM proceeds akin to other Bayesian multimodal FA models. Specifically, FACTM uses $M$ matrices of observations $Y^m \in \mathbb{R}^{N\times D^m}$ (often referred to as views) instead of a single observation matrix $Y$ used by the standard FA. The objective remains to identify common latent factors $Z$ for all of the views, while also deriving view-specific loading matrices $W^m$ \[Y^m = Z (W^m) ' + \varepsilon^m.\]
Following other Bayesian models, we choose the prior for $Z_n$ as $\mathcal{N}(0,I)$, and assume $\varepsilon_{n}$ is normally distributed with a diagonal covariance matrix.  As in \cite{mofa}, in FACTM we assume factor- and feature-wise sparsity on loading matrices $W^m$ using automatic relevance determination and spike-and-slab priors. Namely, each $w_{d, k}^m$ is modeled as the product $\tilde{w}_{d, k}^m \cdot s_{d, k}^m$, and the joint distribution of these two variables is given by
\[p(\tilde{w}_{d, k}^m, s_{d, k}^m) = \mathcal{N}(\tilde{w}_{d, k}^m|0, 1/\alpha_{k}^m)\textrm{Ber}(s_{d, k}^m|\theta_k^m).\]
Additionally, we assume standard conjugate priors on the parameters in the equation above:
$\alpha_k^m \sim \mathcal{G}(a_0^{\alpha}, b_0^{\alpha})$ and $\theta_k^m \sim \textrm{Beta}(a_0^{\theta}, b_0^{\theta})$, and similarly $\varepsilon_{n, d}^m \sim \mathcal{N}(0, 1/\tau_d^m)$, with a conjugate prior on $\tau_d^m$,  $\tau_d^m \sim \mathcal{G}(a_0^{\tau}, b_0^{\tau})$ (all $a_0^{\cdot}, b_0^{\cdot}$ are hyperparameters).


In contrast to other multimodal FA models, FACTM accounts also for {\textit{structured}} data views, where each sample $n$ is represented by a set of $I_n$ data points. Each data point $i$ consists of $J_i$ measured objects $\overline{y}_{n, i, j}$, each with an assigned type $g$, thus it can be represented by a set of count values $\{\overline{y}_{n,i,g}\}_{g=1}^G$, where $\overline{y}_{n, i, g} = \sum_j^{J_i} \mathbb{I}(\overline{y}_{n,i,j} =g)$. Extending the CTM, our model assumes that for each structured view, every sample
(document in the CTM nomenclature) can be represented by a vector of abundances of $L$ clusters (topics). This is modeled by the variable $\eta_n$, which is transformed via the softmax function to yield a probability distribution over the clusters: $\textrm{softmax}(\eta_n) =(\exp(\eta_{n,1})/C, \exp(\eta_{n,2})/C, \ldots, \exp(\eta_{n,L})/C),$ and $C=\sum_{l=1}^{L}\exp(\eta_{n,l}).$ The variable $\eta_n$ is drawn from a multivariate normal distribution $\mathcal{N}_L(\mu_n+\mu^{(0)}, \Sigma^{(0)})$, with $\Sigma^{(0)}$ accounting for the population-level covariance among the clusters. In contrast to standard CTM, we assume that the mean consists of two components: a sample-specific term $\mu_n$, which modifies the population-level variable $\mu^{(0)}$ based on individual characteristics of observation $n$. This sample-specific term constitutes the link between the structured view and the remaining views in the model and depends on the factors $Z$ and view-specific loadings $\overline{W}$ (Fig.~\ref{fig_factm_plate}). Given $\eta_n$, we sample the cluster assignment $\xi_{n,i}$ for each data point (sentence) $i$ from a multinomial distribution: $\textrm{Mult}(1, \textrm{softmax}(\eta_n))$. Each cluster is characterized by a distribution over object types (distinct words), denoted by $g=1,2,\dots, G$. These distributions (topic-word distributions) $\beta_l$ are drawn from a Dirichlet distribution $\textrm{Dir}(\alpha)$. For each observed object (word) $\overline{y}_{n,i,j}$, knowing its cluster assignment $\xi_{n,i}$, we sample its type using the corresponding topic distribution $\textrm{Mult}(1, \beta_{\xi_{n,i}})$. 

The notation used to describe FACTM is summarized in Table D.3 in the Appendix.


\subsection{Inference}
\label{section_model_inference}

The joint probability distribution defining our model (Fig.~\ref{fig_factm_plate}) 
is given by
\begin{subequations}
\begin{align}
&p(Z, W, Y, \overline{W}, \mu, \eta, \xi, \beta, \overline{Y}, \mathcal{X}| \mu^{(0)}, \Sigma^{(0)}, \mathcal{H}) = \nonumber \\
&\quad \prod_{n=1}^N\prod_{k=1}^{K} 
\mathcal{N}(z_{n,k}|0,1) \label{mofa_start} \\ 
&\quad\prod_{m=1}^M\prod_{d=1}^{D_m} \prod_{k=1}^K  \mathcal{N}(\tilde{w}_{d, k}^m|0, 1/\alpha_{k}^m)\textrm{Ber}(s_{d, k}^m|\theta_k^m) \\
&\quad\prod_{m=1}^M \prod_{k=1}^{K} \mathcal{G}(\alpha_k^m|a_0^{\alpha}, b_0^{\alpha}) \textrm{Beta}(\theta_k^m|a_0^{\theta}, b_0^{\theta}) \\
&\quad\prod_{m=1}^M\prod_{n=1}^N\prod_{d=1}^{D_m} \mathcal{N}(y_{n, d}^m|\sum_{k=1}^K z_{n,k} w_{d,k}, 1/\tau_d^m) \\
&\quad\prod_{m=1}^M \prod_{d=1}^{D_m} \mathcal{G}(\tau_d^m|a_0^{\tau}, b_0^{\tau}) \\
&\quad\prod_{k=1}^K \prod_{l=1}^L  \mathcal{N}(\overline{w}_{l,k}|0, 1/\overline{\alpha}_{k})\prod_{k=1}^{K} \mathcal{G}(\overline{\alpha}_k|\overline{a}_0^{\alpha}, \overline{b}_0^{\alpha}) \label{mofa_end}\\
&\quad\prod_{n=1}^N \prod_{l=1}^{L} \mathcal{N}(\mu_{n, l}|\sum_{k=1}^K z_{n,k} \overline{w}_{l,k}, 1/t)  \label{link}\\\
&\quad\prod_{n=1}^N \mathcal{N}_{L}(\eta_n | \mu_{n}+\mu^{(0)}, \Sigma^{(0)}) \label{ctm_start}\\
    &\quad\prod_{n=1}^N \prod_{i=1}^{I_n}\textrm{Mult}(\xi_{n,i}|1,\textrm{softmax}(\eta_n)) \\
   &\quad \prod_{n=1}^N \prod_{i=1}^{I_n} \prod_{j=1}^{J_i}\textrm{Mult}(\overline{y}_{n,i,j}|1,\beta_{\xi_{n,i}})  \prod_{l=1}^L \textrm{Dir}(\beta_l|\alpha_0^{\beta}), \label{ctm_end} 
\end{align}
\end{subequations}
where $\mathcal{X}$ denotes all the remaining nodes in $\{\alpha, \theta, \tau, \overline{\alpha}\}$, and $\mathcal{H}$ all the hyperparameters in $\{a_0^{\cdot}, b_0^{\cdot}, \overline{a}_0^{\alpha}, \overline{b}_0^{\alpha}, t\}$. By default we fix $t=1$, $a_0^{\alpha}=b_0^{\alpha}=\overline{a}_0^{\alpha}=\overline{b}_0^{\alpha} = 1\mathrm{e}{-3}$, $a_0^{\tau}=b_0^{\tau} = 1\mathrm{e}{-3}$ , $a_0^{\theta}=b_0^{\theta} = 1$, $\alpha_0^{\beta}=1$. Note that the
hyperparameters were selected to get non-informative
priors, e.g. for the beta distribution, the parameters
were chosen to make the distribution uniform.

We apply the following mean-field assumption on variational distribution $q$, 
\begin{align}
\label{var_distr}
    &q(Z, W, \overline{W}, \mu, \eta, \xi, \beta, \mathcal{X}) = \prod_{n=1}^N\prod_{k=1}^{K}
q(z_{n,k})  \\ 
&\hspace{0.5em} \prod_{m=1}^M \prod_{k=1}^K  \prod_{d=1}^{D_m}q(\tilde{w}_{d, k}^m,s_{d, k}^m) \prod_{m=1}^M \prod_{k=1}^{K}q(\alpha_k^m)q(\theta_k^m) \nonumber \\
&\hspace{0.5em}\prod_{m=1}^M \prod_{d=1}^{D_m} q(\tau_d^m) \prod_{k=1}^K \prod_{l=1}^L q(\overline{w}_{l,k})\prod_{k=1}^{K} q(\overline{\alpha}_k) \nonumber\\
&\hspace{0.5em} \prod_{n=1}^N q_L(\mu_{n})\prod_{n=1}^N\prod_{l=1}^L q(\eta_{n,l}) \prod_{n=1}^N \prod_{i=1}^{I_n}q(\xi_{n,i}) \prod_{l=1}^L q(\beta_l). \nonumber
\end{align}
Additionally, we assume, that $q_L(\eta_{n})$ follows normal distribution, and $q(\xi_{n,i})$ multinomial distribution.
For the update equations of parameters of variational distributions from lines \eqref{mofa_start}-\eqref{mofa_end} we refer to \cite{mofa}, and for those in \eqref{ctm_start}-\eqref{ctm_end}, we refer to \cite{ctm, ctm_2}.
Additionally, updates for $\mu^{(0)}$ and $\Sigma^{(0)}$ can be found in \cite{ctm_mu_sigma}. Below, we provide update equations for $\mu_n$ from \eqref{link} for $n \in \{1,2,\ldots,N\}$. Note that the distribution $q_L(\mu_n)$ is multivariate. By applying  \eqref{eq_update}, we have
\begin{multline*}
        \log(q_L(\mu_n)) \propto \\ \mathbb{E}_{-\mu_n} \log \Big(\mathcal{N}_L\big(\mu_{n}|\sum_{k=1}^K z_{n,k} \overline{w}_{\cdot,k}, \textrm{diag}(1/t)\big) \\
        \cdot \mathcal{N}_{L}(\eta_n | \mu_{n}+\mu^{(0)}, \Sigma^{(0)})\Big)\\
        =  \mathbb{E}_{-\mu_n} \log \mathcal{N}_{L}(\mu_n|\tilde{\mu}, \tilde{\Sigma})
\end{multline*}
with the parameters 
    $\tilde{\Sigma}^{-1} = \textrm{diag}(t) + \left(\Sigma^{(0)}\right)^{-1}$
    and
    \[\tilde{\mu} = \tilde{\Sigma}\left(\textrm{diag}(t)\sum_{k=1}^K z_{n,k} \overline{w}_{\cdot,k} + \left(\Sigma^{(0)}\right)^{-1}(\eta_n - \mu^{(0)})\right),\]
    which follows from a standard computation for conjugate normal prior on the mean parameter and normal likelihood. Thus, after applying expected value, we get that the updates of the parameters of $q(\mu_n)$ are $\hat \Sigma_{n}^{(\mu)} = \left(\textrm{diag}(t) + \left(\Sigma^{(0)}\right)^{-1}\right)^{-1},$
    \begin{multline*}
        \hat \mu_{n}^{(\mu)} =
        \hat \Sigma_{n}^{(\mu)}\Big(\textrm{diag}(t)\sum_{k=1}^K \langle z_{n,k}\overline{w}_{\cdot,k}\rangle \\+\left(\Sigma^{(0)}\right)^{-1}(\langle\eta_n\rangle - \mu^{(0)})\Big),
    \end{multline*}
    where $\langle\cdot\rangle$ denotes expected value with respect to $q$.
    
The computational complexity of FACTM inference is comparable to the individual components of the model (FA and CTM), as the parameters of the link variable $\mu_n$ are updated in a computationally efficient way, as shown in Figure B.1 in the Appendix.

\subsubsection{Rotations}

We propose a heuristic method for rotating the factors in order to increase their interpretability by associating them with a set of sample features.
We solve this problem using the Kabsch-Umeyama algorithm~\citep{kabsch}. First we compute cross-correlation matrix $H \in \mathbb{R}^{K \times K}$ between $K$ latent factors and $K$ given features. Next, we perform an SVD decomposition on $H$, resulting in $H = USV'$. The rotation matrix is then given by $R = UV'$.

For numerical features, we use Pearson correlation $r$ to compute each element of a matrix $H$. In case of binary features, we also use Pearson correlation $r_{pb}$, specifically known as the point-biserial correlation coefficient, which applies when one of the variables is binary, and the second one continuous. We motivate this choice by the property of $r_{pb}$ that, after a monotonic transformation $r_{pb}((n_0+n_1-2)/(1-r_{pb}^2))^{1/2}$, where $n_0$ denotes the number of $0$s and $n_1$ the number of $1$s in the binary feature, it becomes a test statistic in the unpaired Student's t-test, comparing the means of the two groups.  


\section{EXPERIMENTS}
\label{section_experiments}
We evaluated FACTM on comprehensive simulations,   benchmark datasets and real-world data, measuring performance by: 
(i)  estimation accuracy in versatile scenarios, 
(ii) learning data representations that are predictive in downstream classification tasks, 
(iii) efficacy of the rotation method in obtaining highly interpretable factors, 
(iv) ability to identify meaningful clusters (topics) in the structured data.

We provide code for the experiments and an implementation of FACTM on GitHub\footnote{github.com/szczurek-lab/FACTM}.
\paragraph{Compared methods}  FACTM was compared with several models, which either perform FA on simple views or topic modeling for structured data. For the FA, we used the following models: 
 \textbf{FA Oracle}, serving as the upper bound for FA performance, which replaces the structured views by fixed simple views and fits a standard multimodal FA model. Specifically, for each structured view, we fix the $\mu_n$ variables to their true values (and not infer them  as in FACTM) and represent the structured data as a simple view composed of the fixed $\mu_n$ vectors for the samples $n=1,\ldots,N$;  \textbf{FA+CTM}, which proceeds in two steps: first, it fits CTM model to each structured view and  represents the structured data as a simple view with the estimated $\eta_n - \mu^{(0)}$ as a  feature vector for each sample $n$, and next applies FA to the original and such obtained simple views; and an ablation study that included \textbf{MOFA} (using the implementation from the package by~\cite{muon}), \textbf{muVI} (with the informed version of the model where applicable), \textbf{PCA}~\citep{scikit-learn}, and \textbf{Tucker decomposition}~\citep{tensorly},  which were fitted only to the original simple views (omitting the structured data). For the topic modeling, we compared FACTM with \textbf{CTM}, \textbf{LDA}~\citep{scikit-learn}, and \textbf{ProdLDA} (implemented in the \texttt{pyro} package,~\cite{pyro}). 
\subsection{Parameter estimation accuracy}
First, we evaluated the accuracy of parameter estimation using simulations.
\paragraph{Simulation settings} 
The data were sampled from the model given by Equation (2), with the exception for factor-wise sparsity which was fixed. For the data generation, we used the following settings:: $N=250$ samples, $3$ views (consisting of $2$ observed and $1$ structured), each simple view with $D=10$ features and each structured view with  $L=10$ topics. There were $K=5$ true latent factors, sampled from a standard normal distribution, while the loading matrices were also sampled from standard normal with some columns set to zero (see Fig.~B.2 in the Appendix) to introduce factor-wise sparsity. We applied low feature-wise sparsity, with 10\% of the weights set to zero, resulting in up to 5 elements of $W^m$ being zeroed out. The observations $Y^m$ and the variable $\mu$ were sampled with a variance of $1$. In the structured view, we used $G=100$ distinct words, each sample (document) contained $I=100$ data points (sentences) with $J=10$ objects (words). The population-level variables were set to $\mu_l^{(0)}=0$ for $l=1,2,\ldots, L$, and $\Sigma^{(0)}$ to $5$ on the diagonal and $2.5$ on the upper and lower diagonal, with all other elements set to 0. The topics were sampled from a Dirichlet distribution with equal parameters $\alpha=1$. We devised the following simulation scenarios, each varying one parameter while keeping the rest fixed:
\begin{itemize}
    \item Scenario 1: The link variable $\mu_n$ is multiplied by $\lambda \in \{0,0.5, 1.5, 2\}$. Increasing $\lambda$ strengthens the association of latent factors and structured views.
    \item Scenario 2: The Dirichlet distribution parameter for the topics, $\alpha$, is varied within $\{5,10\}$, with higher values leading to a more uniform topic distribution across words. 
    \item Scenario 3: The number of topics is changed to $L\in\{5, 15\}$.
    \item Scenario 4: The variable $\mu^{(0)}$ is multiplied by $\lambda_{\mu^{(0)}}$, for $\lambda_{\mu^{(0)}} \in \{0.25, 0.5, 0.75, 1\}$. A higher $\lambda_{\mu^{(0)}}$ value results in greater disproportion in the baseline topic proportions. Here, $\mu^{(0)}$ is a centered, log-transformed vector scaled to sum to 1 of equally distributed values between 1 and 3.
    \item Scenario 5: The covariance matrix $\Sigma^{(0)}$ is scaled by $\lambda_{\Sigma^{(0)}} \in \{0.2, 0.6\}$.
\end{itemize}
The basic set of parameters is obtained by fixing $\lambda=1$, $\alpha=1$, $L=10$, $\lambda_{\mu^{(0)}}=0$, $\lambda_{\Sigma^{(0)}}=1$. 

\begin{figure}
\includegraphics[width=0.48\textwidth]{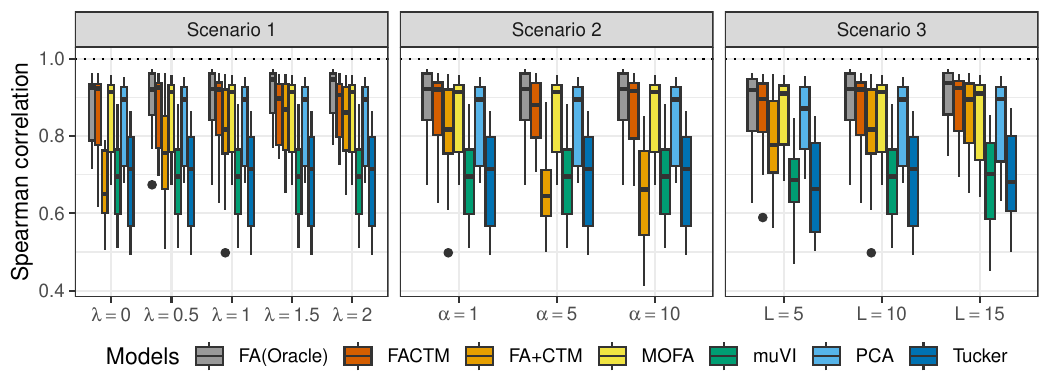}
\caption{Comparison of true factors and optimally reordered latent factors in factor analysis models.}
\label{fig_sim_fa}
\end{figure}

\begin{figure}[ht]
\includegraphics[width=0.48\textwidth]{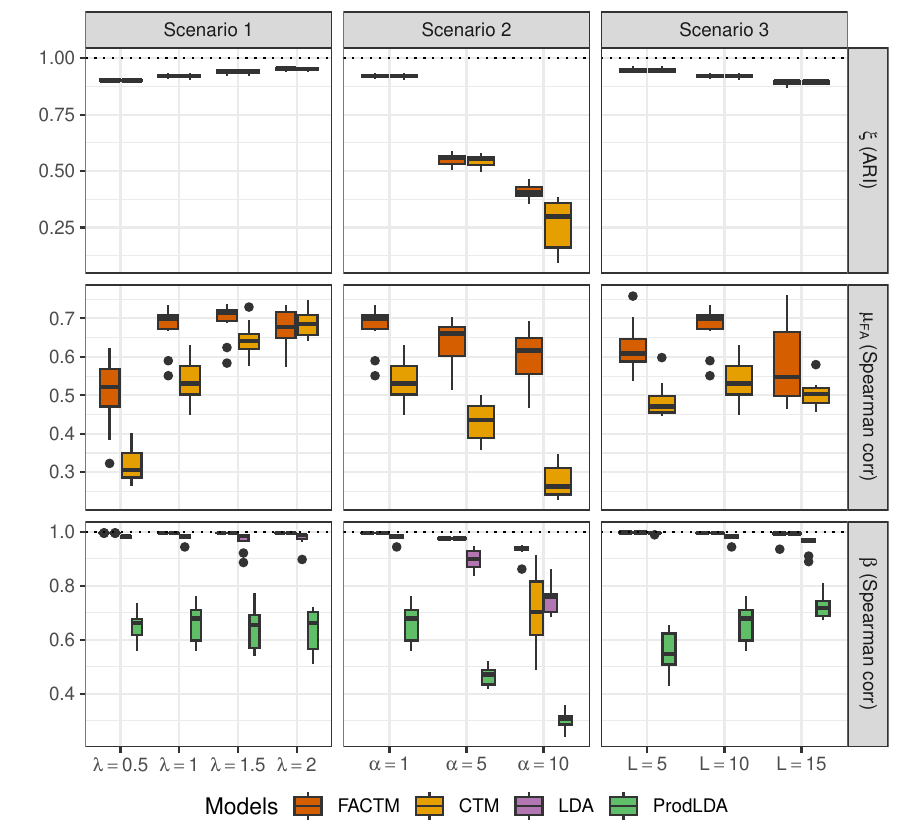}
\caption{Comparison of true parameters and inferred parameters following the optimal reordering of topics in topic models.}
\label{fig_sim_ctm_v1}
\end{figure}

\begin{figure}[ht]
\includegraphics[width=0.48\textwidth]{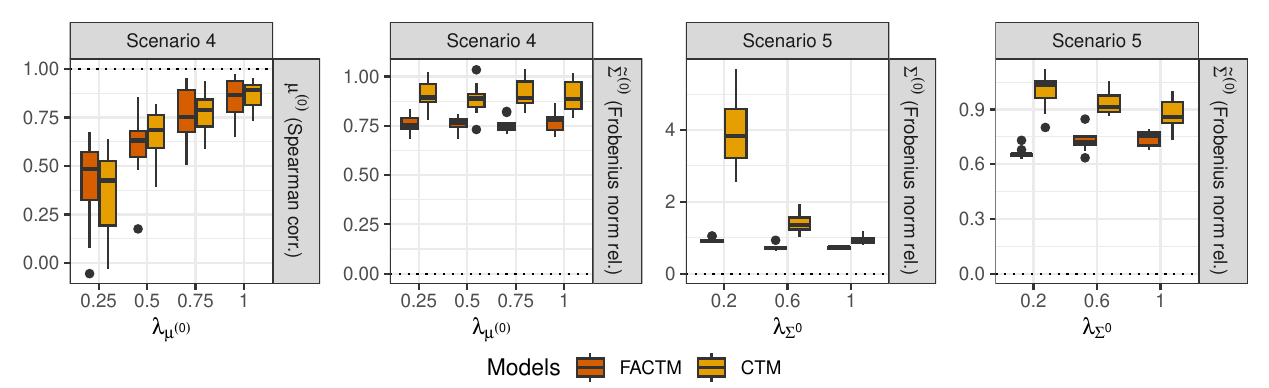}
\caption{Comparison of true and inferred population-level variables in FACTM and CTM (other topic models do not account for population mean and covariance of topics). The Frobenius distance is computed relative to the Frobenius norm of the true covariance matrix. $\tilde{\Sigma}^{(0)}$ represents the covariance matrix scaled to have ones on the diagonal. Dashed lines indicate optimal performance.}
\label{fig_sim_ctm_v2}
\end{figure}

\paragraph{Evaluation} In the simulations, we first evaluated the 
inferred latent factors by comparing them to the ground truth. Given the permutation invariance of factors, we first computed Spearman correlation for all the estimated-true factor pairs and used the Hungarian method~\citep{hungarian} to best match the inferred factors with the true ones (Fig.~\ref{fig_sim_fa}).
We also evaluated the structured view part of the model by assessing how accurately it estimated the parameters of the latent variables compared to the true parameters, specifically focusing on the performance of the link variable $\mu_n$ 
(for the CTM model, we use $\eta_n - \mu^{(0)}$ as a substitute for the link variable $\mu_n$, which is not explicitly estimated in that model),
clustering variable $\xi$, topic distribution $\beta$ (Fig.~\ref{fig_sim_ctm_v1}), and population-level variables $\mu^{(0)}$ and $\Sigma^{(0)}$ (Fig.~\ref{fig_sim_ctm_v2}). Since topic order is also not-identifiable, we applied the Hungarian method to the contingency table of true versus inferred topic assignments to determine the optimal ordering. Each simulation was repeated $10$ times.

\paragraph{Results}  
FACTM is more accurate in factor estimation compared to other models across all simulated scenarios (Fig.~\ref{fig_sim_fa}). 
For some settings, the additional information transferred from structured part modeled by CTM to FA in FA+CTM model actually deteriorates the fit, emphasizing the importance of joint estimation as performed by FACTM (Fig.~\ref{fig_sim_fa} for $\lambda=0$ or $\alpha = 10$). 
Interestingly, for simulations, simple PCA performs relatively well, as the sparsity is not excessively high, while muVI performs suboptimally. 
In more sparse scenarios, however,  PCA's performance deteriorates since it does not account for sparsity, whereas muVI shows even better performance than MOFA (Fig.~B.3).  Figure~B.4 further illustrates the importance of incorporating sparsity-inducing priors in FACTM.

FACTM also outperforms other methods in estimating parameters for structured data (Fig.~\ref{fig_sim_ctm_v1}, \ref{fig_sim_ctm_v2}). This advantage is particularly pronounced for the inference of the covariance matrix $\Sigma^{(0)}$, as seen in Figure~\ref{fig_sim_ctm_v2}. Figure.~B.5 compares  $\Sigma^{(0)}$ estimated by FACTM and CTM, showing that, unlike FACTM, CTM fails to capture the true structure of $\Sigma^{(0)}$.

\subsection{Predictive power of learned latent representations}
We next evaluated the hidden representations inferred by the models on two benchmarks, and one real-world music dataset.

\paragraph{Benchmark datasets}
We used two multimodal datasets of opinion video clips as benchmarks, with each clip labeled by its sentiment: CMU-MOSI~\citep{mosi} and CMU-MOSEI~\citep{mosei}. To form simple views 
we averaged the vision and audio measurements across time stamps,  obtaining a single vector per view per sample. 
For the structured view, we used the transcriptions. Due to the short length of the texts, in FACTM each sentence consisted of a single word for these datasets. 
 \paragraph{Real-world dataset: Mirex} Mirex is a dataset containing songs~\citep{mirex_data}. For this dataset, we extracted four simple views: the first containing standard acoustic features extracted using the \texttt{pyAudioAnalysis} package~\citep{pyaudioanalysis}, the second with melody-based features (melody was extracted using the Melodia vamp plug-in~\citep{melodia_salamon} with features obtained as in~\cite{salamon_features}), the third  with features extracted using the \texttt{essentia} package~\citep{essentia}, and the fourth with common text features like average word length. All the views were quantile normalized. For the structured view, we used the song lyrics, treating each line as a sentence. Each song is labeled with one of five classes: class 1 (described as boisterous, confident, passionate, rousing, rowdy), class 2 (amiable/good natured, cheerful, fun, rollicking, sweet), class 3 (autumnal, bittersweet, brooding, literate, poignant, wistful), class 4 (campy, humorous, silly, witty, wry, whimsical), and class 5 (agressive, fiery, intense, tense/anxious, visceral, volatile). Note that given the descriptions, class 3 is clearly negative, while class 4 is positive. We also retrieved the genre of the songs.
\paragraph{Evaluation} 
To assess the hidden representations learned by the models, we evaluated how informative they were in relation to the assigned labels. 
Specifically, as sample representations we used either latent factors (Tab.~\ref{tab_bench_fa}), or representations derived from the topics: $\mu_n$ for FACTM, $\eta_n - \mu^{(0)}$ for CTM, and log-transformed probabilities for LDA (Tab.~\ref{tab_bench_ctm}).  For each dataset, we used 10-fold cross-validation to train random forest to predict sample labels using these representations as input features. The classification performance was measured using the AUC for the ROC and Precision-Recall curves (for PR-AUC, see Tab.~C.1 and~C.2). In the case of CMU-MOSI and CMU-MOSEI, the sentiments were binarized. 
For the multi-label Mirex dataset, we applied a one-versus-rest approach and calculated the weighted average performance across the classes. 

   
\paragraph{Results} Classification  on benchmark datasets' representations confirms the simulation results. For the factor-based representations, differences between models are minimal (Tab.~\ref{tab_bench_fa}). 
For the structured data, FACTM performs best in two cases and consistently outperforms the ablation model - CTM (Tab.~\ref{tab_bench_ctm}).

\begin{table}[ht]
\caption{Performance of the random forest classifier using inferred latent factors across the datasets. The mean ROC-AUC $\pm$ standard deviation values over 10-fold cross-validation are reported.} \label{tab_bench_fa}
\begin{center}
\begin{tabular}{l|lll}
  \hline
 & Mirex & Mosei & Mosi \\ 
  \hline
  FACTM & $0.67\pm 0.02$ & $\textbf{0.73}\pm 0.01$ & $\textbf{0.68}\pm 0.04$ \\ 
  FA+CTM & $\textbf{0.68}\pm 0.04$ & $\textbf{0.73}\pm 0.01$ & $0.67\pm 0.04$ \\ 
 MOFA & $\textbf{0.68}\pm 0.03$ & $\textbf{0.73}\pm 0.01$ & $0.66\pm 0.04$ \\ 
 muVI & $\textbf{0.68}\pm 0.02$ & $0.70\pm 0.01$ & $0.67\pm 0.04$ \\ 
   \hline
\end{tabular}
\end{center}
\end{table}
\begin{table}
\caption{Performance of the random forest classifier using model-specific sample representations across the datasets. The mean ROC-AUC $\pm$ standard deviation values over 10-fold cross-validation are reported.}\label{tab_bench_ctm}
\begin{center}
\begin{tabular}{l|lll}
  \hline
  & Mirex & Mosei & Mosi \\ 
  \hline
   FACTM & $\textbf{0.64}\pm 0.06$ & $0.63\pm 0.01$ & $\textbf{0.61}\pm 0.06$ \\ 
  CTM & $0.57\pm 0.04$ & $0.55\pm 0.02$ & $0.57\pm 0.05$ \\ 
   LDA & $0.62\pm 0.05$ & $\textbf{0.66}\pm 0.01$ & $0.57\pm 0.04$ \\ 
   \hline
\end{tabular}
\end{center}
\end{table}

\begin{figure}
\includegraphics[width=0.48\textwidth]{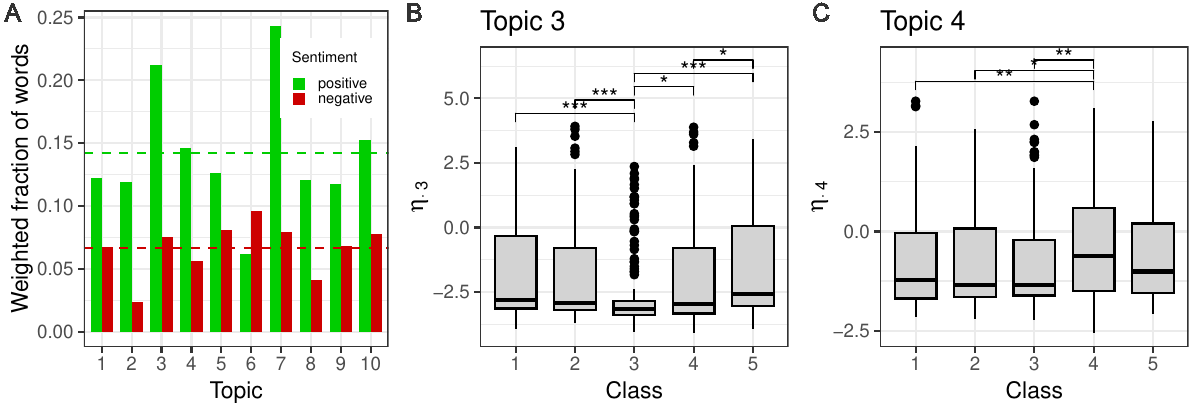}
\caption{\textbf{A}. Topic's average positivity and negativity, measured as the weighted average of positive/negative words in a topic. \textbf{B}\&\textbf{C}. The values of $\eta_{\cdot, 3}$ (\textbf{B}) and $\eta_{\cdot, 4}$ (\textbf{C}) split by class membership of the samples with two-sided  Wilcoxon pairwise tests. 
Stars denote significance of Bonferroni-adjusted p-values: \raisebox{-0.6ex}{***} $< 0.001$, \raisebox{-0.6ex}{**} $< 0.01$, \raisebox{-0.6ex}{*} $< 0.05$.}
\label{fig_topic_sentiment}
\end{figure}

Since for the Mirex data, the topic prevalences quantified by FACTM-inferred $\eta_n$ provided the most predictive representations for class labels, we further explored the insights offered by this representation. 
Topic~3 shows the second-highest abundance of positive words (Fig.~\ref{fig_topic_sentiment}A). At the same time, the probability of the lyrics lines (sentences) in songs (samples) from the negative class~3 being assigned to this topic is lower compared to samples from other classes (Fig.~\ref{fig_topic_sentiment}B). Conversely, sentences in samples from the positive class~4 have a higher probability of being clustered in Topic~4 (in which the number of positive words exceeds the average and negative words are below average, see Fig.~\ref{fig_topic_sentiment}A and~\ref{fig_topic_sentiment}C). For graphical representation of 
word frequencies across all topics, see wordclouds (Fig.~B.6).

\subsection{Enhancement of interpretability using the rotation method}
Further, we showcased the rotation method and the factor interpretability that it brings using the the music dataset Mirex as the testbed.

\begin{figure}
\includegraphics[width=0.48\textwidth]{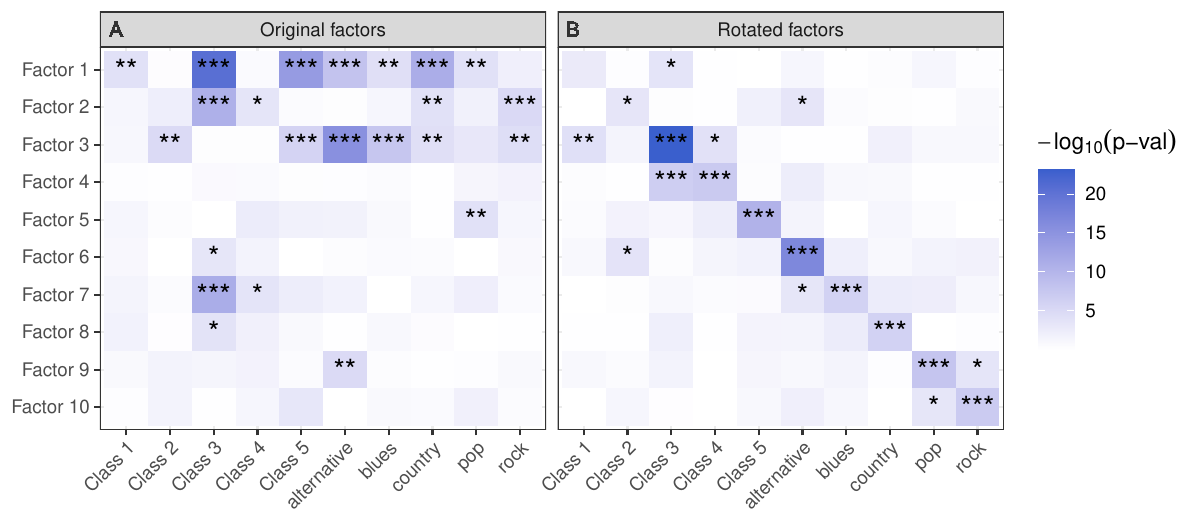}
\caption{Log-transformed p-values of the two-sided Wilcoxon test for 
latent factors and binary features (\textbf{A}) compared to rotated latent factors and binary features (\textbf{B}). 
Stars denote significance of Bonferroni-adjusted p-values: \raisebox{-0.6ex}{***} $< 0.001$, \raisebox{-0.6ex}{**} $< 0.01$, \raisebox{-0.6ex}{*} $< 0.05$.
}
\label{fig_rotations}
\end{figure}

\paragraph{Results} 
To showcase the rotation method, we compared the associations of the originally inferred factors with binary features of the samples in the Mirex dataset to the associations of the rotated factors.
For the original factors (Fig.~\ref{fig_rotations}A) the binary features have an association mainly with three first factors. After applying our rotation method we obtained many assosiations on the diagonal, improving the interpretability of the consecutive factors (Fig.~\ref{fig_rotations}B). Additionally, values of the rotated factors better discriminate between the five classes than the original factors (Fig.~B.7). We further show the interpretability of the most discriminative rotated factors by listing their top-weighted features (Fig.~B.8).

\subsubsection{Ability to infer meaningful topics in real-world, multimodal COVID-19 data}
Finally, we demonstrated applicability of FACTM to large scale, real-world multi-modal biological data and its ability to retrieve meaningful clustering in the structured data. 
\paragraph{Real-world COVID-19 dataset} The dataset consisted of multi-modal data from long COVID-19 patients~\citep{long_covid_published} and included the following modalities: flow cytometry from bronchoalveolar lavage (BAL) fluids, two time-separated sets of CT scans, and scRNA-seq data, which included expression measurements for many single cells per sample. Thus, the dataset consisted of three simple views: one flow cytometry-based view with general cell type fractions per sample, and two CT-based views with the fractions of scans occupied by radiographic abnormalities per sample, as well as one structured view, with each cell treated as a sentence, where gene counts were interpreted as word counts.  All simple views underwent quantile normalization, while the structured view was count-normalized, and 1000 highly variable genes were selected for the analysis. We used 20 samples, for which the data for all the views was available, resulting in 169,741 single cells included in the analysis.
\paragraph{Evaluation} For this dataset, our main objective was to cluster single cells from the scRNA-seq data and compare FACTM clusters with cell types obtained by scVI method, followed by manual curation, as described in the original article. 

\begin{figure}[ht]
\includegraphics[width=0.48\textwidth]{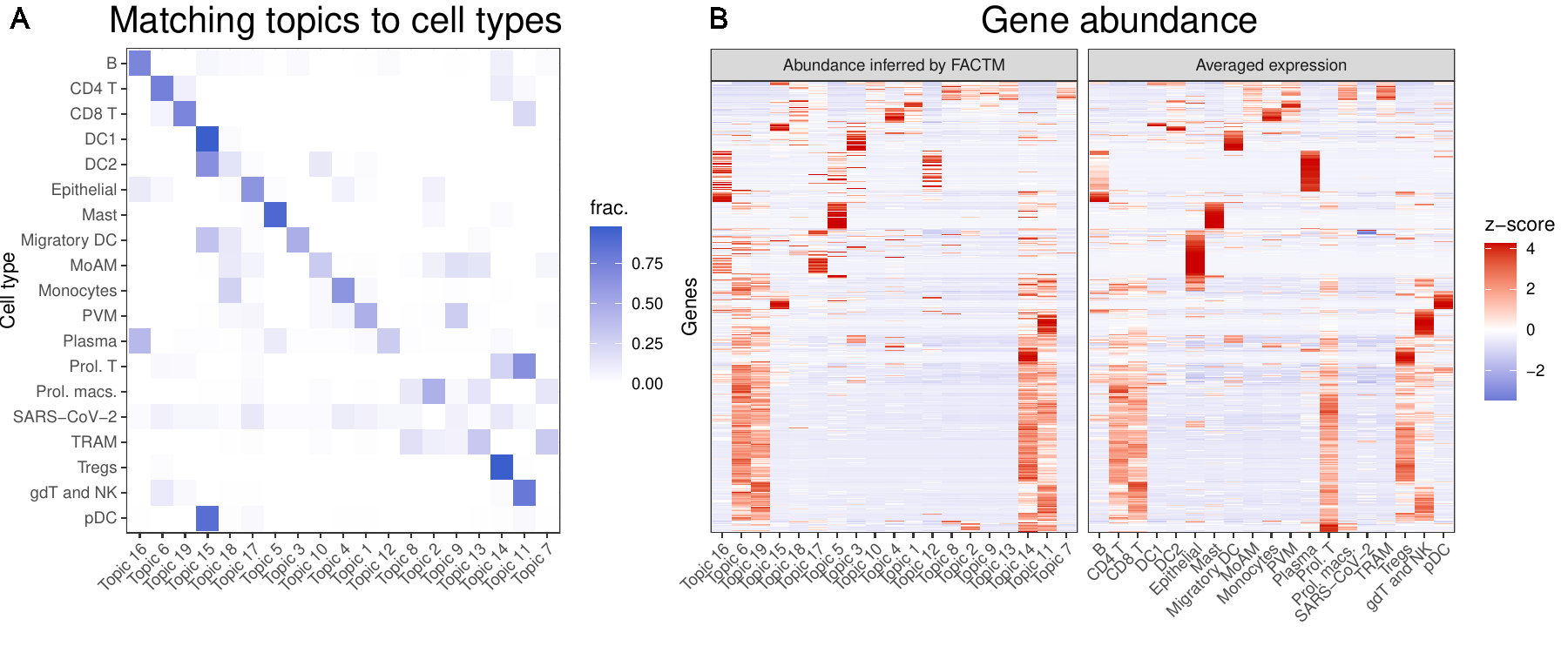}
\caption{\textbf{A}. Contingency table of cell types and clusters inferred by FACTM, scaled such that rows sum to 1. The topics are ordered based on the application of the Hungarian method. \textbf{B}. Comparison of z-scored true and inferred average gene expressions for each cell type and each topic. The ordering of the genes in the two panels is the same.}
\label{fig_covid}
\end{figure}

\paragraph{Results} 
 FACTM  identifies biologically relevant clusters in real-world scRNA-seq data  (Fig.~\ref{fig_covid}). Indeed, FACTM 
 groups cells into the same cell types as identified in a dedicated clustering and expert annotation step in the original article (Fig.~\ref{fig_covid}A). Even in cases, when one cluster covers more than one cell type, the grouping remains meaningful, e.g. Topic~15 contains cells classified as DC1, DC2, Migratory DC or pDC in the original paper, all of which are subtypes of dendritic cells. Similarly, Topic~16 groups B cells and plasma cells, often referred to as plasma B cells (Fig.~B.9). This biological relevance is further supported by the correspondence between gene expression profiles within clusters found by FACTM and the respective cell types, indicating that FACTM learned the correct topic distributions  (Fig.~\ref{fig_covid}B).

\section{LIMITATIONS AND FUTURE WORK}

The proposed model could be further extended in several ways. First, although most hyperparameters in our approach were
  chosen to result in non-informative priors, or were
  fixed across all methods for consistent comparison, this
  may be seen as a limitation of our current study. In
  future work, we aim to enhance our hyperparameter selection process by incorporating automatic learning techniques, such as the prior predictive matching
  method proposed in \cite{prior_learning}. Secondly, FACTM assumes linear dependencies between the views and latent factors, and the analytical solutions used for optimization in variational inference rely on the conjugacy of parametric distributions, which limits the model's expressiveness. To address this issue, we plan to extend our model to incorporate nonlinear dependencies.

\section{CONCLUSIONS}
In this work, we addressed the challenge of modeling multi-view and multi-structured data by proposing FACTM, a novel probabilistic Bayesian method that integrates factor analysis with correlated topic model  and uses variational inference for optimization. In extensive simulations, FACTM achieved superior accuracy in inferring factors, topics, and their covariance structure. Both factor- and topic-based sample representations learned by FACTM showed state-of-the-art predictive power for label classification on benchmark and real-world music data. On the COVID-19 data, FACTM found meaningful biological clusters. The proposed supervised factor rotation method enhanced factor interpretability for the music dataset. In fact, the rotation method can be used to
improve interpretability of any latent factor model. In summary, 
FACTM adapts the established FA method to handle the complex, structured data of the current world.


\subsubsection*{Acknowledgements}
This work received funding from the Polish National Science Centre SONATA BIS grant No. 2020/38/E/NZ2/00305.

\bibliography{paper_final}

\section*{Checklist}



 \begin{enumerate}

 \item For all models and algorithms presented, check if you include:
 \begin{enumerate}
   \item A clear description of the mathematical setting, assumptions, algorithm, and/or model. [\underline{Yes}/No/Not Applicable] \\
   \textit{See Section \ref{section_model_description} for model description and Section \ref{section_model_inference} for optimization details.}
   \item An analysis of the properties and complexity (time, space, sample size) of any algorithm. [\underline{Yes}/No/Not Applicable] \\
   \textit{We compare computational complexity of our model to existing methods in Section \ref{section_model_inference}.}
   \item (Optional) Anonymized source code, with specification of all dependencies, including external libraries. [\underline{Yes}/No/Not Applicable] \\
   \textit{We will provide a link to the GitHub repository after the reviewing process is complete and the anonymity requirements are lifted.}
 \end{enumerate}

 \item For any theoretical claim, check if you include:
 \begin{enumerate}
   \item Statements of the full set of assumptions of all theoretical results. [\underline{Yes}/No/Not Applicable]
   \item Complete proofs of all theoretical results. [\underline{Yes}/No/Not Applicable]
   \item Clear explanations of any assumptions. [\underline{Yes}/No/Not Applicable]     
 \end{enumerate}

 \item For all figures and tables that present empirical results, check if you include:
 \begin{enumerate}
   \item The code, data, and instructions needed to reproduce the main experimental results (either in the supplemental material or as a URL). [\underline{Yes}/No/Not Applicable] \\
   \textit{The code is available via the GitHub repository, which will be made accessible as soon as the anonymity requirements are lifted.}
   \item All the training details (e.g., data splits, hyperparameters, how they were chosen). [\underline{Yes}/No/Not Applicable]
    \item A clear definition of the specific measure or statistics and error bars (e.g., with respect to the random seed after running experiments multiple times). [\underline{Yes}/No/Not Applicable]
    \item A description of the computing infrastructure used. (e.g., type of GPUs, internal cluster, or cloud provider). [Yes/\underline{No}/Not Applicable]
 \end{enumerate}

 \item If you are using existing assets (e.g., code, data, models) or curating/releasing new assets, check if you include:
 \begin{enumerate}
   \item Citations of the creator If your work uses existing assets. [\underline{Yes}/No/Not Applicable]
   \item The license information of the assets, if applicable. [\underline{Yes}/No/Not Applicable] \\
   \textit{See Section \ref{section_experiments}, paragraph on 'Compared Methods'.}
   \textit{We will include the licence information in the GitHub repository.}
   \item New assets either in the supplemental material or as a URL, if applicable. [\underline{Yes}/No/Not Applicable] \\
   \textit{We will upload the code to the GitHub repository, which will be made accessible as soon as the anonymity requirements are lifted.}
   \item Information about consent from data providers/curators. [Yes/No/\underline{Not Applicable}] \\
   \textit{We use publicly available datasets.}
   \item Discussion of sensible content if applicable, e.g., personally identifiable information or offensive content. [Yes/No/\underline{Not Applicable}]
 \end{enumerate}

 \item If you used crowdsourcing or conducted research with human subjects, check if you include:
 \begin{enumerate}
   \item The full text of instructions given to participants and screenshots. [Yes/No/\underline{Not Applicable}]
   \item Descriptions of potential participant risks, with links to Institutional Review Board (IRB) approvals if applicable. [Yes/No/\underline{Not Applicable}]
   \item The estimated hourly wage paid to participants and the total amount spent on participant compensation. [Yes/No/\underline{Not Applicable}]
 \end{enumerate}

 \end{enumerate}

\onecolumn
\appendix
\setcounter{figure}{0}
\renewcommand{\thefigure}{B.\arabic{figure}}

\section{DERIVATIONS}
 FACTM combines factor analysis (FA) with correlated topic model (CTM), and while many of the variational update equations follow the ones for the original models (specifically, \cite{mofa} for the FA and \cite{ctm} for the CTM part), our extension required modifications and additional updates.

In Section \ref{section_updates_ctm}, we present the update formulas of variational parameters in coordinate ascent algorithm for the variable $\mu_n$, which links the FA and CTM parts of the model, as well as for the  CTM adapted for sentence-based structured data. While the latter represents a relatively small extension to the standard CTM \citep{ctm}, we provide the equations for completeness. The key difference lies in the fact that sentences introduce a constraint, where each data point (sentence), composed of a set of observed objects (words), must originate from a single topic. The original CTM becomes a special case of our model, where there is one word per sentence. On the one hand, this special case is most flexible, allowing for a different distribution of topics for each observed object (word). On the other hand, our extension accounts for a reasonable assumption that each entire sentence groups words of the same topic. 
Moreover, in some applications assuming multi-object data point is crucial for capturing the nature of the data (see e.g. COVID-19 example in the main text, where sentences correspond to single cells, words to measured genes and the topics correspond to cell types).

Section \ref{section_fa_remarks}  addresses the adjustments in the update formulas for the FA part in FACTM.

\subsection{Variational inference for the structured model based on CTM with sentences}
\label{section_updates_ctm}

\paragraph{Overview of notation and assumptions} Recall from the main text, that the ELBO is defined as 
\begin{equation}
\label{eq_elbo_def}
    ELBO(q) := \mathbb{E}_q\log p(X,Y) - \mathbb{E}_q\log q(X) = \mathbb{E}_q\log p(X,Y) + H(q).
\end{equation}
In this equation, $q$ represents a variational distribution, while $X$ and $Y$ hidden and observed variables, respectively. The  term $-\mathbb{E}_q\log q(X)$ in \eqref{eq_elbo_def} corresponds to the entropy denoted by $H$ of the distribution $q$. 
The optimal variational distribution $q_i \vcentcolon = q(X_i)$ is determined by the expression:
\begin{equation}
\label{eq_elbo_e}
    \log q(X_i) \propto \E_{-q_i} \log p(X,Y),
\end{equation}
where the expectation is taken over all variational distributions except $q_i$ corresponding to the variable $X_i$. Alternatively, assuming a parametric form for the variational distribution, the ELBO can also be optimized directly.
 
 We simplify the joint distribution of the FACTM model showed in Figure 1 from the main text by focusing only on the structured component. For this consideration, we assume that the nodes related to the factor analysis part, $z_{n,k}$ and $\overline{w}_{l,k}$ for $n \in \{1,2,\ldots,N\}$, $k\in\{1,2,\ldots,K\}$, $l \in \{1,2,\ldots,L\}$, are fixed (note that this simplification does not affect the correctness of the update equations in the full model). Thus, based on Equation (2) from the main text, the joint probability density we consider is given by
\begin{align*}
p(\overline{W}, \mu, \eta, \xi, \beta, \overline{Y}| \mu^{(0)}, \Sigma^{(0)}, t, \alpha_0^{\beta}) &= 
   \prod_{n=1}^N \prod_{l=1}^{L} \mathcal{N}(\mu_{n, l}|\sum_{k=1}^K z_{n,k} \overline{w}_{l,k}, 1/t)   \prod_{n=1}^N \mathcal{N}_{L}(\eta_n | \mu_{n}+\mu^{(0)}, \Sigma^{(0)}) \\
    &\quad
    \prod_{n=1}^N \prod_{i=1}^I\textrm{Mult}(\xi_{n,i}|1,\textrm{softmax}(\eta_n)) \prod_{l=1}^L \textrm{Dir}(\beta_l|\alpha_0^{\beta})  \\ 
    & \quad\prod_{n=1}^N \prod_{i=1}^I \prod_{j=1}^{J_i}\textrm{Mult}(\overline{y}_{n,i,j}|1,\beta_{\xi_{n,i}}) 
\end{align*}
where
$\textrm{softmax}(\eta_n) = (\exp(\eta_{n,1})/C, \exp(\eta_{n,2})/C, \ldots, \exp(\eta_{n,L})/C)$ and $C=\sum_{l=1}^{L}\exp(\eta_{n,l}).$ In this scenario, the variational distribution, assuming a mean-field approximation along with a parametric form for $q(\eta_{n,l})$ and $q(\xi_{n,i})$ as in the main text (see Equation (3) there),  takes the following form, incorporating the variational parameters:
\begin{equation*}
    q(\mu, \eta, \xi, \beta) =  \prod_{n=1}^N q_L(\mu_n|\mu_{n}^{(\mu)},\Sigma_{n}^{(\mu)})\prod_{n=1}^N\prod_{l=1}^L \mathcal{N}(\eta_{n,l}|\mu_{n, l}^{(\eta)}, (\sigma_{n, l}^{(\eta)})^2) \prod_{n=1}^N \prod_{i=1}^{I_n}\textrm{Mult}(\xi_{n,i}|1, \phi_{n,i})\prod_{l=1}^L q(\beta_l|\alpha_{l}^{(\beta)}).
\end{equation*}
  Thus a formula for the ELBO for this specific model is given by
\begin{subequations}
\begin{align}
    ELBO(q) = &\sum_{n=1}^N \Big(\sum_{l=1}^L\E_{q} \log\mathcal{N}(\mu_{n,l}|\sum_{k=1}^{}z_{n,k}\overline{w}_{l,k}, 1/t)\label{eq:kl_line1}\\
    &\qquad+ \E_{q}\log\mathcal{N}_{L}(\eta_n | \mu_{n}+\mu^{(0)}, \Sigma^{(0)}) \label{eq:kl_line2}\\
    &\qquad+ \sum_{i=1}^{I_n} \E_{q}\log\textrm{Mult}(\xi_{n,i}|1,\textrm{softmax}(\eta_n)) \label{eq:kl_line3}\\
    &\qquad+ \sum_{i=1}^{I_n} \sum_{j=1}^{J_i} \E_{q}\log \textrm{Mult}(\overline{y}_{n,i,j}|1,\beta_{\xi_{n,i}}) \Big) \label{eq:kl_line4} \\
    & +\sum_{l=1}^L \E_q\log\textrm{Dir}(\beta_l|\alpha)\label{eq:kl_line5}\\
&+H(q_L(\mu_{n}))+H(q(\eta_{n,l}))+H(q(\xi_{n,i}))+ H(q(\beta_{l})).
\label{eq:kl_line6}
\end{align}
\end{subequations}
The entropy of $q$ in \eqref{eq:kl_line6} for the variables, for which the parametric distribution is assumed, equals
\begin{equation}
\label{eq:entropy_eta}
    H(q(\eta_{n,l})) = \log((\sigma_{n,l}^{(\eta)})^2)/2 + \log(\sqrt{2\pi e})
\end{equation}
and
\begin{equation}
\label{eq:entropy_xi}
H(q(\xi_{n,i})) = \sum_{l=1}^L \phi_{n, i, l}\log(\phi_{n, i, l}).
\end{equation}
Now we obtain the update equations for all the variational parameters, all of which are listed below, including the parameters for the link variable $\mu_n$. Additionally, we provide also updates for $\mu^{(0)}$ and $\Sigma^{(0)}$.

\paragraph{Variational parameters in the structered part of the model}
\begin{itemize}
\item $\mu_n$ for $n=1,2,\ldots, N$\\
     Variational parameters: $\mu_{n}^{(\mu)}$, $\Sigma_{n}^{(\mu)}$
    \item $\eta_n$ for $n=1,2,\ldots, N$\\
     Variational parameters: $\mu_{n}^{(\eta)}$, $(\sigma_{n}^{(\eta)})^2$, and additional parameter $\zeta_n$ introduced in Section \ref{section_updates}
     \item $\xi_{n,i}$ for $n=1,2,\ldots,N$, $i=1,2,\ldots,I_n$\\
     Variational parameters: $\phi_{n,i}$
     \item $\beta_l$ for $l=1,2,\ldots,L$\\
     Variational parameters: $\alpha_{l}^{(\beta)}$
\end{itemize}

\subsubsection{The update equations}
\label{section_updates}

For clarity in the following sections, we introduce the notation $\tilde{\xi}_{n,i}$. Recall that $\xi_{n,i}$ is clustering variable that takes a value from the set of topics $\{1,2,\ldots,L\}$. The variable $\tilde{\xi}_{n,i}$ is a one-hot encoded vector, where all elements are zero except for the $l$th position, which is set to 1 if $\xi_{n,i} = l$.

\begin{itemize}
  \item $\mu_n$  for $n=1,2,\ldots, N$\\
We provide a detailed derivation of the updates for $\mu_{n}^{(\mu)}$ and $\Sigma_{n}^{(\mu)}$ from Section 3.3 of the main text. Using~\eqref{eq_elbo_e}, rather than optimizing the ELBO directly, we can compute the following expected value:
    \begin{equation*}
        \log(q(\mu_n)) \propto  \mathbb{E}_{q} \log \big(\mathcal{N}_L(\mu_{n}|\sum_{k=1}^K z_{n,k} \overline{w}_{\cdot,k}, \textrm{diag}(1/t)) 
        \cdot \mathcal{N}_{L}(\eta_n | \mu_{n}+\mu^{(0)}, \Sigma^{(0)})\big),
\end{equation*}
where $\mathbb{E}_{-\mu_n}$ denotes the expectation taken over all variables following the distribution  $q$, except for $\mu_n$. If we interpret the second normal density as $\mathcal{N}_{L}(\eta_n - \mu^{(0)} | \mu_{n}, \Sigma^{(0)})$, and treat the first normal distribution $\mathcal{N}_L(\mu_{n}|\sum_{k=1}^K z_{n,k} \overline{w}_{\cdot,k}, \textrm{diag}(1/t)) $ as the prior for $\mu_n$ and the second $\mathcal{N}_{L}(\eta_n - \mu^{(0)} | \mu_{n}, \Sigma^{(0)})$ as the likelihood, then the posterior distribution is also normal $\mathcal{N}_{L}(\mu_n|\tilde{\mu}, \tilde{\Sigma})$, with parameters given by
 \[\tilde{\Sigma} = \left(\textrm{diag}(t) + \left(\Sigma^{(0)}\right)^{-1}\right)^{-1}\]
    and
    \[\tilde{\mu} = \tilde{\Sigma}\left(\textrm{diag}(t)\sum_{k=1}^K z_{n,k} \overline{w}_{\cdot,k} + \left(\Sigma^{(0)}\right)^{-1}(\eta_n - \mu^{(0)})\right).\]
     Thus, after applying expected value, we get that the updates of the parameters of $q_L(\mu_n)$ are 
     \[\hat \Sigma_{n}^{(\mu)} = \left(\textrm{diag}(t) + \left(\Sigma^{(0)}\right)^{-1}\right)^{-1},\]
    \begin{equation*}
        \hat \mu_{n}^{(\mu)} =
        \hat 
 \Sigma_{n}^{(\mu)}\Big(\textrm{diag}(t)\sum_{k=1}^K z_{n,k}\overline{w}_{\cdot,k} +\left(\Sigma^{(0)}\right)^{-1}(\E_q\eta_n - \mu^{(0)})\Big),
    \end{equation*}
    where $\E_q \eta_n = \mu_n^{(\eta)}$. We recall that for now, focusing solely on the structured part, we have assumed that both $z_{n,k}$ and $\overline{w}_{\cdot,k}$ are deterministic. However, in the complete FACTM, these are treated as random variables, and their expectations must also be computed.
     \item $\eta_n$  for $n=1,2,\ldots, N$\\
     As our objective is to maximize the $ELBO$ in equations \eqref{eq:kl_line1} - \eqref{eq:kl_line6} with respect to variational parameters of $\eta_n$, the relevant terms involving $\eta_n$ are \eqref{eq:kl_line2}, \eqref{eq:kl_line3}, and the corresponding entropy term in \eqref{eq:kl_line6}. Since direct optimization is infeasible, we derive a lower bound as in \cite{ctm} (up to constant terms with respect to parameters $\mu_{n}^{(\eta)}$ and $(\sigma_{n}^{(\eta)})^2$) by introducing a new auxiliary parameter $\zeta_n > 0$. Since there is no analytic solution for maximizing the lower bound with respect to $\mu_{n}^{(\eta)}$ and $(\sigma_{n}^{(\eta)})^2$, we compute the gradient and employ the L-BFGS-B algorithm for optimization. \\
     \vspace{0.1cm} \\
     First, we focus on the term \eqref{eq:kl_line3}. This gives
   \begin{equation}
    \label{eq:eta_non_conjugate}
        \E_{q}\log\textrm{Mult}(\xi_{n,i}|1,\textrm{softmax}(\eta_n)) \propto \E_{q} \log \left(\frac{\exp(\eta_{n}' \tilde{\xi}_{n,i})}{\sum_{l=1}^{L} \exp(\eta_{n,l})}\right) 
        =  \E_{q}\eta_{n}' \tilde{\xi}_{n,i} - \E_{q} \log \left(\sum_{l=1}^{L} \exp(\eta_{n,l})\right),
    \end{equation}
    where $\propto$ indicates equality up to terms constant with respect to $\mu_{n}^{(\eta)}$ and $(\sigma_{n}^{(\eta)})^2$.  
    Next, we apply a Taylor expansion of the logarithm around the auxiliary parameter $\zeta_n$
    \[ \log \left(\sum_{l=1}^{L} \exp(\eta_{n,l})\right) = \log(\zeta_n) + \sum_{m=1}^{\infty} \frac{(-1)^{m-1}}{m\zeta_n^m}\left(\sum_{l=1}^{L} \exp(\eta_{n,l}) - \zeta_n\right)^m\]
    to upper bound the second term of Equation \eqref{eq:eta_non_conjugate}
    \[\E_{q} \log \left(\sum_{l=1}^{L}  \exp(\eta_{n,l})\right) \leq \log(\zeta_n) + \zeta_n^{-1}\sum_{l=1}^{L} \E_q \exp(\eta_{n,l}) -1.\]
   Summing over $i\in\{1,2,\ldots,I_n\}$, we lower bound term  \eqref{eq:kl_line3} in the following way
    \begin{equation}
        \label{eq:mult_eta_E_q}
        \sum_{i=1}^I\E_{q}\log\textrm{Mult}(\tilde{\xi}_{n,i}|1,\textrm{softmax}(\eta_n)) \geq \sum_{i=1}^I \E_{q} \eta_{n}' \tilde{\xi}_{n,i} - I\big(\log(\zeta_n) + \zeta_n^{-1}\sum_{l=1}^{L} \E_{q}\exp(\eta_{n,l}) - 1\big).
    \end{equation}
    The expected values in \eqref{eq:mult_eta_E_q} equal
 \begin{equation}
        \label{eq:mult_eta_E_q_1}
    \E_{q} \eta_{n}' \tilde{\xi}_{n,i} = \sum_{l=1}^L\mu_{{n,l}}^{(\eta)}\phi_{n,i,l}
    \end{equation}
    and
    \begin{equation}
        \label{eq:mult_eta_E_q_2}
        \E_{q}\exp(\eta_{n,l}) = \exp\left(\mu_{n,l}^{(\eta)} + (\sigma_{{n,l}}^{(\eta)})^2/2\right),
        \end{equation}
        where we use the fact, that the second expectation is the moment generating function of a normal distribution evaluated at 1.
    Next, we expand the term \eqref{eq:kl_line2}
    \begin{equation}
        \label{eq:normal_eta_E_q}
        \E_{q}\log\mathcal{N}_{l}(\eta_n | \mu_{n}+\mu^{(0)}, \Sigma^{(0)}) \propto 
        - \E_{q}(\eta_n - \mu_n - \mu^{(0)})'\left(\Sigma^{(0)}\right)^{-1}(\eta_n - \mu_n - \mu^{(0)})/2,
    \end{equation}
    and compute the expected value of a quadratic form
    \begin{multline}
     \label{eq:normal_eta_E_q_exp}
        \E_{q}(\eta_n - \mu_n - \mu^{(0)})'\left(\Sigma^{(0)}\right)^{-1}(\eta_n - \mu_n - \mu^{(0)}) = \textrm{tr}\left(\left(\textrm{diag}((\sigma_{n}^{(\eta)})^2) + \Sigma_n^{(\mu)}\right)\left(\Sigma^{(0)}\right)^{-1}\right) \\ + (\mu_{n}^{(\eta)} - \mu_{n}^{(\mu)} - \mu^{(0)})'\left(\Sigma^{(0)}\right)^{-1}(\mu_{n}^{(\eta)} - \mu_{n}^{(\mu)} - \mu^{(0)}),
    \end{multline}
    where $\textrm{diag}((\sigma_{n}^{(\eta)})^2)$ denotes an $L\times L$ diagonal matrix with $(\sigma_{n,l}^{(\eta)})^2$ as its diagonal elements.
     Thus, by combining the results from \eqref{eq:kl_line2}, \eqref{eq:kl_line3}, and the entropy from \eqref{eq:entropy_eta}, we get the lower bound that needs to be maximized with respect to $(\mu_{n}^{(\eta)}, (\sigma_{n}^{(\eta)})^2)$, and $\zeta_n$:
    \begin{multline*}
        f(\mu_{n}^{(\eta)}, (\sigma_{n}^{(\eta)})^2, \zeta_n) = 
        -\textrm{tr}\left(\textrm{diag}\left((\sigma_{n}^{(\eta)})^2\right)\left(\Sigma^{(0)}\right)^{-1}\right)/2 - (\mu_{n}^{(\eta)} - \mu_{n}^{(\mu)} - \mu^{(0)})'\left(\Sigma^{(0)}\right)^{-1}(\mu_{n}^{(\eta)} - \mu_{n}^{(\mu)} - \mu^{(0)})/2 \\
        + \sum_{l=1}^L\mu_{{n,l}}^{(\eta)}\sum_{i=1}^I\phi_{n,i,l} - I\left(\log(\zeta_n) + \zeta_n^{-1}\sum_{l=1}^{L}  \exp\left(\mu_{n,l}^{(\eta)} + (\sigma_{{n,l}}^{(\eta)})^2/2\right)\right) 
        +\sum_{l=1}^{L}\log((\sigma_{n, l}^{(\eta)})^2)/2.
    \end{multline*}
Gradients of $f$ equal
        \begin{align*}
        &\nabla_{\mu_{n}^{(\eta)}} f(\mu_{n}^{(\eta)} ,(\sigma_{n}^{(\eta)})^2, \zeta_n) = 
        - \left(\Sigma^{(0)}\right)^{-1}(\mu_{n}^{(\eta)} - \mu_{n}^{(\mu)} - \mu^{(0)}) + \sum_{i=1}^I \phi_{n,i}  - I\zeta_n^{-1} \exp\left(\mu_{n}^{(\eta)} + (\sigma_{n}^{(\eta)})^2/2\right), \\
        &\nabla_{(\sigma_{n}^{(\eta)})^2} f(\mu_{n}^{(\eta)} ,(\sigma_{n}^{(\eta)})^2, \zeta_n) = 
        -\textrm{diag}\left(\left(\Sigma^{(0)}\right)^{-1}\right)/2 -  I\zeta_n^{-1} \exp\left(\mu_{n}^{(\eta)} + (\sigma_{n}^{(\eta)})^2/2\right)/2  + 1/(2(\sigma_{n}^{(\eta)})^2).
    \end{align*}
    All terms in the equations above are vectors of length $L$, and the transformations are applied elementwise (e.g. $\exp(\mu_{n}^{(\eta)}) = (\exp(\mu_{n,1}^{(\eta)}), \exp(\mu_{n,2}^{(\eta)}), \ldots, \exp(\mu_{n,L}^{(\eta)}) $). The optimal value of $\zeta_n$ obtained analytically from $f$ equals
    \[\hat \zeta_n = \sum_{l=1}^{L}  \exp(\mu_{n,l}^{(\eta)} + (\sigma_{{n,l}}^{(\eta)})^2/2).\]
    \item $\xi_{n,i}$ for $n=1,2,\ldots,N$, $i=1,2,\ldots,I_n$ \\
    As the terms of ELBO involving $\xi_{n,i}$ are \eqref{eq:kl_line3}, \eqref{eq:kl_line4} and the corresponding entropy term in \eqref{eq:kl_line6} (computed in \eqref{eq:entropy_xi}), for fixed $n, i, l$ the function we aim to maximize with respect to $\phi_{n,i,l}$ up to constants is
    \begin{align*}
         f(\phi_{n,i,l}) &= \E_{q} \eta_{n,l} \tilde{\xi}_{n,i,l} + \sum_{j=1}^{J_i}\E_q \left(\tilde{\xi}_{n,i,l}\log \beta_{l, y_{n, i, j}}\right) - \phi_{n, i, l} \log \phi_{n, i, l}\\
    &=\mu_{{n,l}}^{(\eta)}\phi_{n,i,l} + \phi_{n, i, l} \sum_{j=1}^{J_i}\E_{q}\log \beta_{l, y_{n, i, j}} - \phi_{n, i, l} \log \phi_{n, i, l}.
    \end{align*}
  This results in
    \[\hat \phi_{n, i, l} \propto \exp\left(\mu_{{n,l}}^{(\eta)}+\sum_{j=1}^{J_i}\E_{q}\log \beta_{l, \tilde{y}_{n, i, j}}\right),\]
    where $\hat \phi_{n, i, l}$ is known up to multiplicative constants. That issue is solved by noting that $\sum_{l=1}^L\phi_{n, i, l} = 1$,  and after rescaling, we obtain the final value for  $\hat  \phi_{n, i, l}$. \\
    Now we compute the term $\sum_{j=1}^{J_i}\E_{q}\log \beta_{l, \tilde{y}_{n, i, j}}$. Note that
    \[ \sum_{j=1}^{J_i}\E_{q}\log \beta_{l, \nu_{n, i, j}} = \sum_{g=1}^G \underbrace{\sum_{j=1}^{J_i}\I(\overline{y}_{n,i,j}=g) }_{\substack{\textrm{number of words }g\\ \textrm{ in a sentence }i}} \E_{q} \log \beta_{l, g} = \sum_{g=1}^G \overline{y}_{n,i,g}\E_{q} \log \beta_{l, g}.\]
    As a variational distribution of $\beta_l$ is $\textrm{Dirichlet}(\alpha_l^{(\beta)})$, we have that $\beta_{l, g_0}$ is $\textrm{Beta}(\alpha_{l,g_0}^{(\beta)}, \sum_{g\neq g_0}\alpha_{l,g}^{(\beta)})$ and hence we obtain
    \begin{equation}
         \label{eq:logbeta}
   \E_{q}\log \beta_{l, g_0} = \psi(\alpha_{l, g_0}^{(\beta)}) - \psi \left(\sum_{g=1}^G\alpha_{l, g}^{(\beta)}\right),
    \end{equation}
    where $\psi$ is digamma function.
    \item $\beta_l$ for $l=1,2,\ldots,L$\\
    For $\beta_l$, we assume conjugate distributions, where the prior is a Dirichlet distribution and the likelihood follows a Multinomial distribution, thus we simply have that the optimal $q(\beta_l)$ is also Dirichlet with parameters
    \[\alpha_{l,g}^{(\beta)} = \alpha_0^{\beta} + \sum_{n=1}^N \sum_{i=1}^{I_n} \phi_{n, i, l} \overline{y}_{n,i,g}.\]
    \item $\mu^{(0)}$ and $\Sigma^{(0)}$\\
To get updates of $\mu^{(0)}$ and $\Sigma^{(0)}$, we need to maximize a function (see \eqref{eq:normal_eta_E_q} and \eqref{eq:normal_eta_E_q_exp})
     \begin{multline*}
        \sum_{n=1}^{N} \Big(-\log(\textrm{det}(\Sigma^{(0)}))/2
         -\textrm{tr}\left(\left(\textrm{diag}((\sigma_{n}^{(\eta)})^2) + \Sigma_n^{(\mu)}\right)\left(\Sigma^{(0)}\right)^{-1}\right)/2 \\ - (\mu_{n}^{(\eta)} - \mu_{n}^{(\mu)} - \mu^{(0)})'\left(\Sigma^{(0)}\right)^{-1}(\mu_{n}^{(\eta)} - \mu_{n}^{(\mu)} - \mu^{(0)})/2 \Big) \\
        =\sum_{n=1}^{N} \Big(-\log(\textrm{det}(\Sigma^{(0)}))/2
        -\textrm{tr}\big(\big(\textrm{diag}((\sigma_{n}^{(\eta)})^2) + \Sigma_n^{(\mu)}) \\ +(\mu_{n}^{(\eta)} - \mu_{n}^{(\mu)} - \mu^{(0)})(\mu_{n}^{(\eta)} - \mu_{n}^{(\mu)} - \mu^{(0)})'\big)\left(\Sigma^{(0)}\right)^{-1}\big)/2 \Big).
    \end{multline*}
    This task is equivalent to finding ML estimators of the parameters of multivariate normal distribution, thus
    \[\mu^{(0)} = \frac{1}{n}\sum_{n=1}^N \left(\mu_{n}^{(\eta)} - \mu_{n}^{(\mu)}\right),\]
    \[\Sigma^{(0)} = \frac{1}{n} \sum_{n=1}^N \left(\Sigma^{(\mu)} +  \textrm{diag}((\sigma_{n}^{(\eta)})^2) +(\mu_{n}^{(\eta)} - \mu_{n}^{(\mu)} - \mu^{(0)})(\mu_{n}^{(\eta)} - \mu_{n}^{(\mu)} - \mu^{(0)})'\right).\]
\end{itemize}

\paragraph{Remark} It is important to note that the update equations do not require the counts $\overline{y}_{i,j,g}$ to be integers. Instead, they can also be interpreted as weights, as long as they remain positive.

\subsection{Variational inference for the FA part}
\label{section_fa_remarks}

For the FA part, we provide only the final update equations for $\overline{w}_{l,k}$ and $z_{n,k}$, as their derivations follow standard procedures, such as those in \cite{mofa}, and can be easily obtained from Equation (1) from the main text.  The key modification is that unobserved $y_{n,l}$ is replaced by the expectation of the link variable $\mu_{n,l}$, namely $\E_q \mu_{n,l}=\mu_{n,l}^{(\mu)}$. Additionally, the data precision $\tau$ is substituted by parameter $t$ for structured views, and we do not use spike-and-slab prior for loadings $\overline{w}$ in these structured view.

\paragraph{Selected variational parameters in FA part of the model}
\begin{itemize}
\item $\overline{w}_{l,k}$ for $l=1,2,\ldots, L$, $k=1,2,\ldots, K$\\
     Variational parameters: $\mu_{l,k}^{(\overline{w})}$, $(\sigma_{l,k}^{(\overline{w})})^2$
    \item $z_{n,k}$ for $n=1,2,\ldots, N$, $k=1,2,\ldots, K$\\
     Variational parameters: $\mu_{n,k}^{(z)}$, $(\sigma_{n,k}^{(z)})^2$
\end{itemize}

\subsection{The update equations}

\begin{itemize}
    \item $\overline{w}_{l,k}$ for $l=1,2,\ldots, L$, $k=1,2,\ldots, K$\\
    The variational distribution of $\overline{w}_{l,k}$ is a normal distribtuion and the updates for $\mu_{l,k}^{(\overline{w})}$ and $(\sigma_{l,k}^{(\overline{w})})^2$ are
    \begin{align*}
        \hat \mu_{l,k}^{(\overline{w})} &= t(\hat \sigma_{l,k}^{(\overline{w})})^{2}\left(\sum_{n=1}^{N}\E_q z_{n,k} \left( \E_q\mu_{n,l} - \sum_{k'\neq k} \E_q \overline{w}_{l,k'}\E_q z_{n,k'}\right)\right),\\
        (\hat \sigma_{l,k}^{(\overline{w})})^2 &= \left(t\sum_{n=1}^N \E_q z_{n,k}^2 + \E_q \overline{\alpha}_k\right)^{-1},
    \end{align*}
    where $\E_q \mu_{n,l} = \mu_{n,l}^{(\mu)}$, $\E_q \overline{w}_{l,k} = \mu_{l,k}^{(\overline{w})}$, and the remaining expected values are analogous to those in \cite{mofa}.
    \item $z_{n,k}$ for $n=1,2,\ldots, N$, $k=1,2,\ldots, K$\\
    The variational distribution of $z_{n,k}$ a normal distribution and the updates for $\mu_{n,k}^{(z)}$ and $(\sigma_{n,k}^{(z)})^2$ are
    \begin{align*}
        \hat \mu_{n,k}^{(z)} &= (\hat \sigma_{n,k}^{(z)})^{2}\Biggl(\sum_{m=1}^{M}\sum_{d=1}^{D_m} \E_q \tau_d^m \E_q w_{d,k}^m  \biggl(y_{n,d}^m - \sum_{k'\neq k}\E_q z_{n,k'} \E_q w_{d,k'}^m\biggl) \\
        & \qquad \sum_{l=1}^L t\E_q \overline{w}_{l,k}\biggl(\E_q \mu_{n,l} - \sum_{k'\neq k}\E_q z_{n,k'} \E_q \overline{w}_{l,k'}\biggl)\Biggr),\\
        (\hat \sigma_{n,k}^{(z)})^2 &= \left(1 + \sum_{m=1}^M \sum_{d=1}^{D_m} \E_q (w_{d,k}^m)^2 + \sum_{l=1}^{L} \E_q (\overline{w}_{l,k})^2\right)^{-1},
    \end{align*}
    where $\E_q \mu_{n,l} = \mu_{n,l}^{(\mu)}$, $\E_q \overline{w}_{l,k} = \mu_{l,k}^{(\overline{w})}$, $\E_q (w_{d,k}^m)^2 = ( \mu_{l,k}^{(\overline{w})})^2 + ( \sigma_{l,k}^{(\overline{w})})^2$, and the remaining expected values are analogous to those in \cite{mofa}.
\end{itemize}
\vfill
\newpage 
\section{ADDITIONAL FIGURES}

\begin{figure}[h!]
    \centering
    \includegraphics[width=0.85\linewidth]{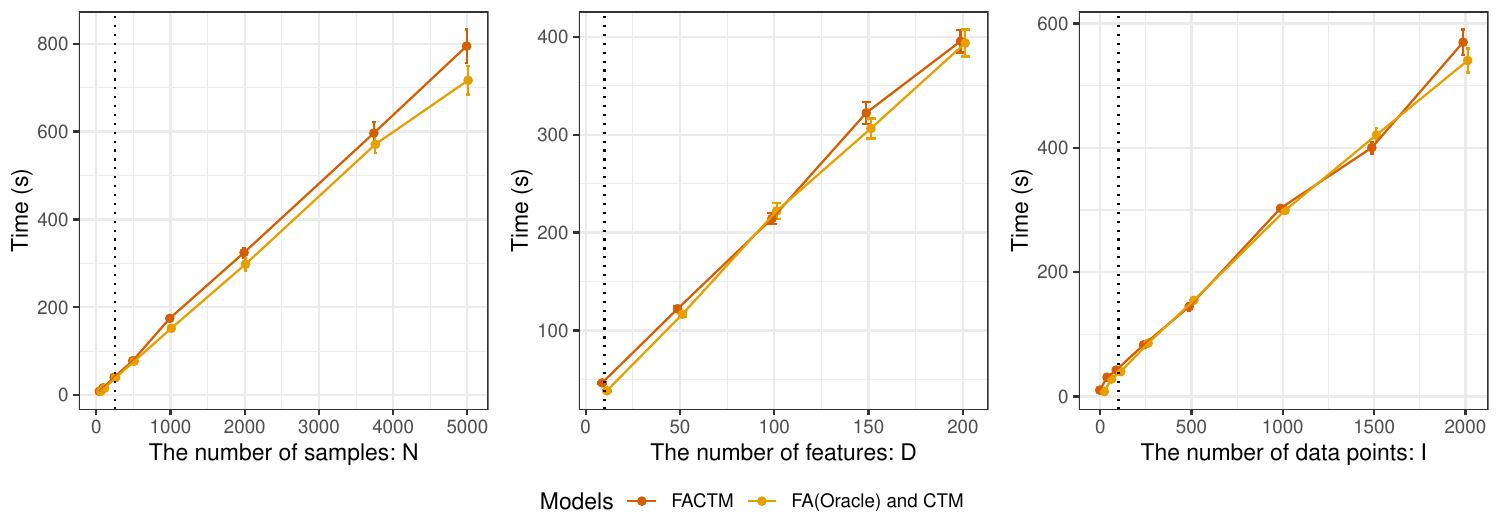}
    \caption{Runtime analysis using the baseline simulation setup (marked with vertical lines) showing scalability dependence on key data parameters: $N$, $D$, and $I$. The execution time of FACTM is compared to the combined runtime of its individual components, FA (FA Oracle) and CTM, over 10 iterations of each algorithm across 5 model fits on 10 datasets. For each dataset, the results of the 5 models were averaged. Data points represent these averages, with error bars showing the standard deviation across datasets.}
    \label{fig:enter-label}
\end{figure}

\begin{figure}[h!]
\centering
\includegraphics[width=0.5\textwidth]{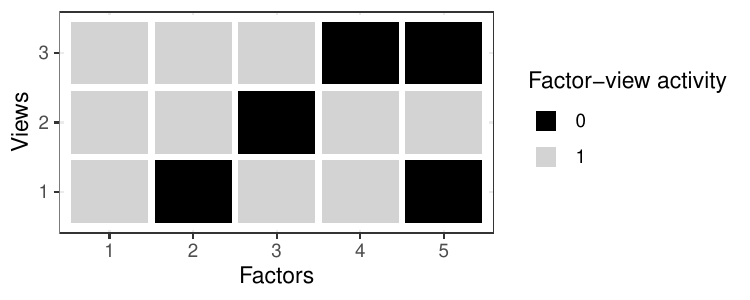}
\caption{Illustration of factor-wise sparsity used in simulations, where 0 indicates an inactive factor for a given view, and 1 indicates an active factor.}
\label{sup_fig_sim_plot}
\end{figure}

\begin{figure}[h!]
\centering
\includegraphics[width=0.45\textwidth]{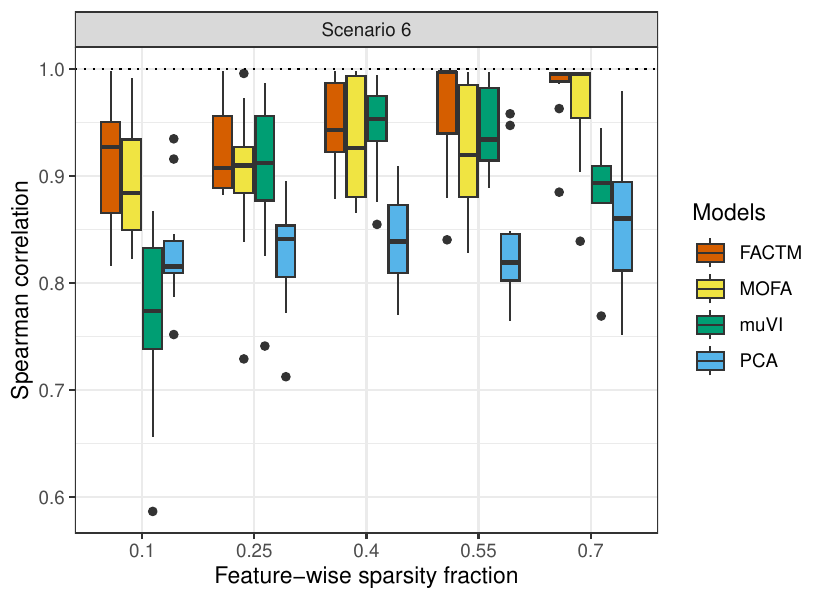}
\caption{Comparison of true factors and optimally
reordered latent factors for muVI, MOFA and PCA in an additional simulation scenario (Scenario 6), which explores varying levels of feature-wise sparsity.\\
\textbf{Scenario 6:} This scenario is based on our standard setup, with a few modifications. We increased the number of features in both simple views to $D=500$ (to support the inference of sparsity-related parameters for both models). The basic feature-wise sparsity fraction, initially set at $0.1$, was increased to four values: $\{0.25, 0.4, 0.55, 0.7\}$.}
\label{sup_fig_sim_muvi}
\end{figure}

\begin{figure}[h!]
\centering
\includegraphics[width=0.45\textwidth]{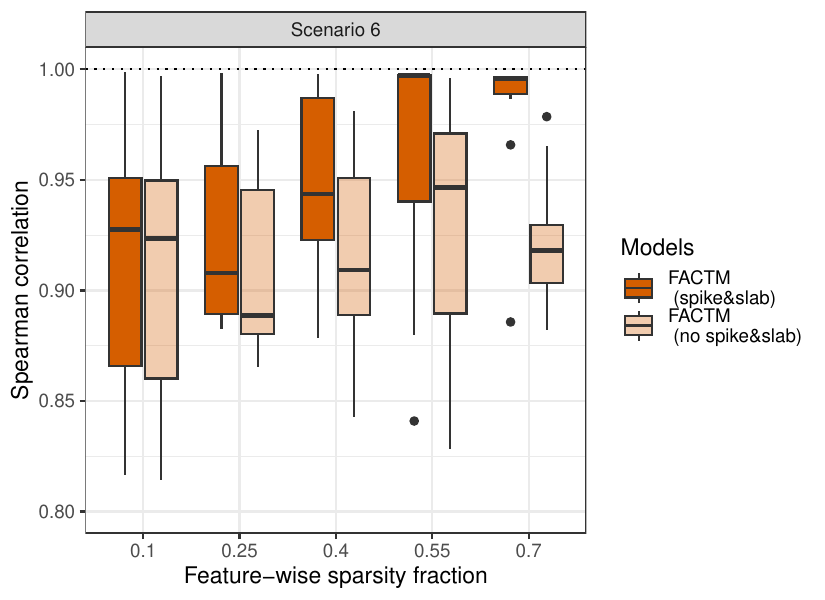}
\caption{Comparison of true factors and optimally
reordered latent factors for FACTM with and without spike and slab prior in an additional simulation scenario (Scenario 6), which explores varying levels of feature-wise sparsity.\\
For the description of Scenario 6 see Figure B.3.}
\label{sup_fig_sim_spike}
\end{figure}

\begin{figure}[h!]
\centering
\includegraphics[width=0.6\textwidth]{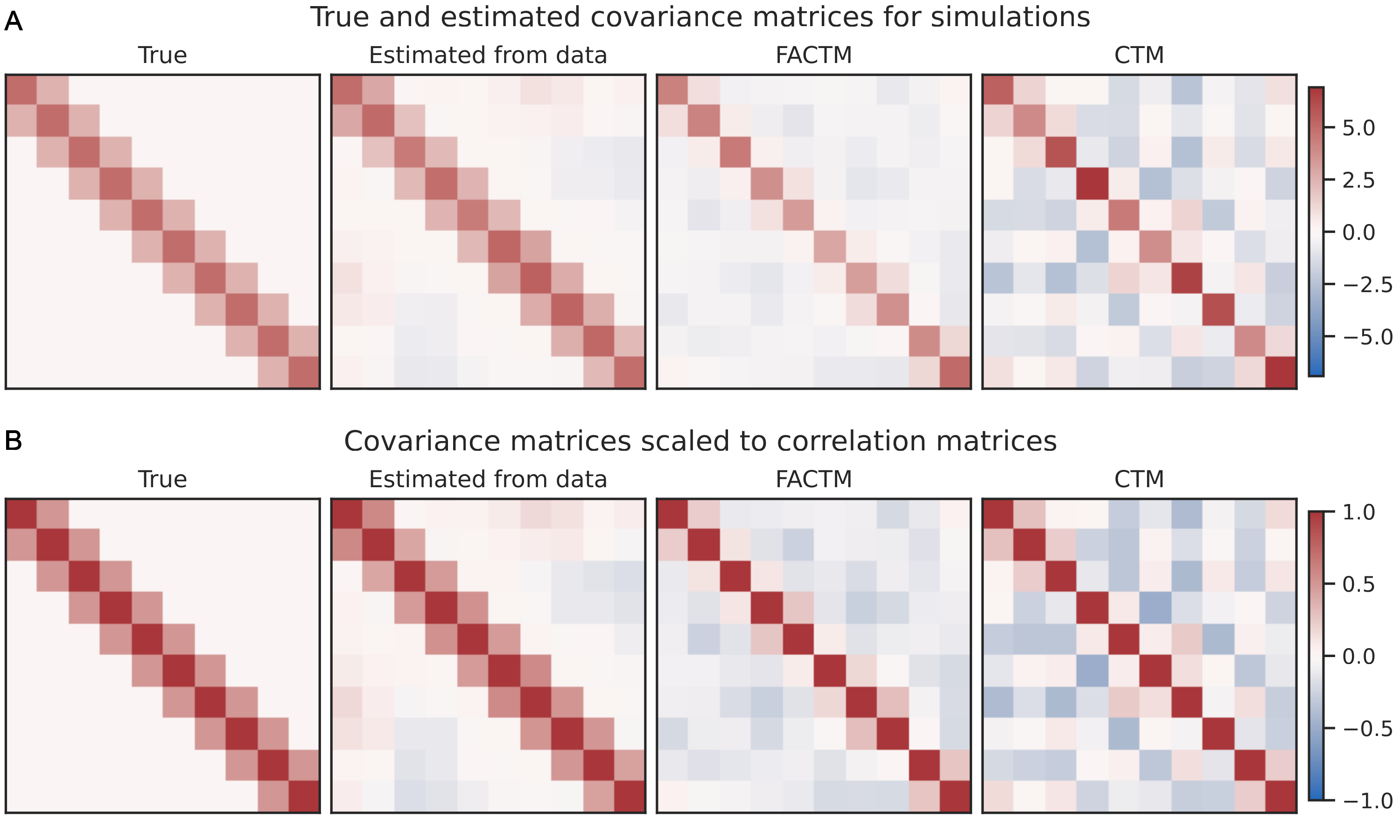}
\caption{Comparison of true ($\Sigma^{(0)}$) and estimated covariance matrices (\textbf{A}) and covariance matrices scaled to have 1 on the diagonal (\textbf{B}). From left to right: the true matrix used in the simulations; a matrix estimated using the true values of $\mu_n$, $\eta_n$, and $\mu^{(0)}$; the matrix inferred by FACTM; and the matrix inferred by CTM.}
\label{sub_fig_cov}
\end{figure}

\begin{figure}[h!]
\centering
\includegraphics[width=0.9\textwidth]{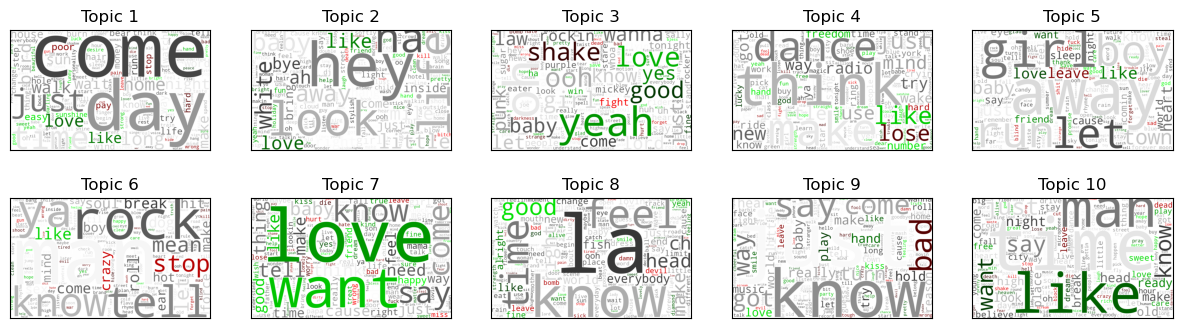}
\caption{Wordclouds representing the topics inferred by FACTM for the Mirex dataset. Words associated with positive sentiment are displayed in green, those with negative sentiment are shown in red,  while the neutral words are gray. The size of each word corresponds to the probability of its occurrence within the topic.}
\label{fig_after_rot}
\end{figure}

\begin{figure}[h!]
\centering
\includegraphics[width=0.9\textwidth]{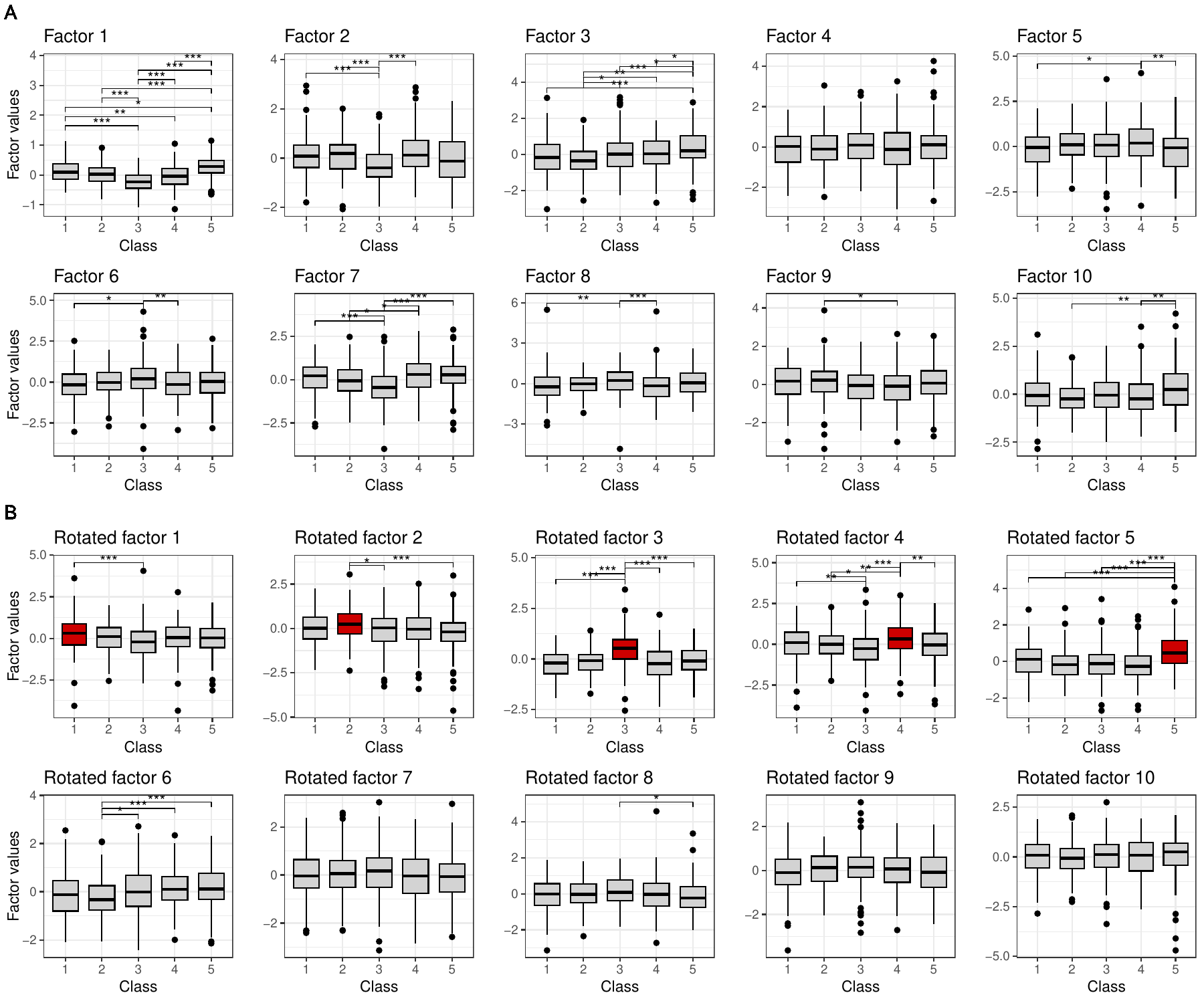}
\caption{The latent factors (\textbf{A}) and rotated latent factors (\textbf{B}) split by class membership of the samples.  In red, we highlight the class and rotated factor pairs expected to be associated with each other by the construction of the rotation method. Significance of the differences between factor value distributions in the classes were evaluated with two-sided  Wilcoxon pairwise tests.
Stars denote significance of Bonferroni-adjusted p-values: \raisebox{-0.6ex}{***} $< 0.001$, \raisebox{-0.6ex}{**} $< 0.01$, \raisebox{-0.6ex}{*} $< 0.05$.}
\label{sup_fig_factors_classes}
\end{figure}


\begin{figure}[h!]
\centering
\includegraphics[width=1\textwidth]{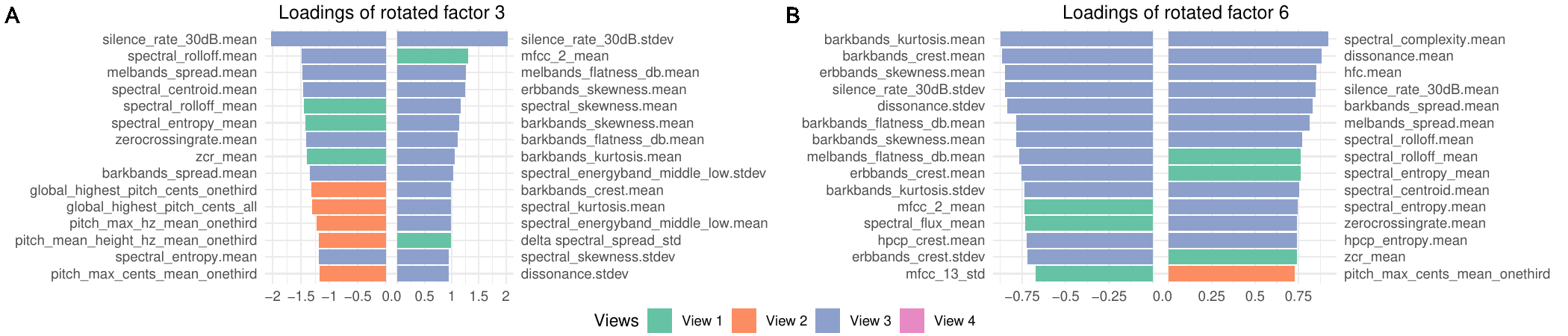}
\caption{Top-weighted features of simple views for rotated factor 3 (\textbf{A}) and rotated factor 6 (\textbf{B}), colored by the view to which they belong. Since the input data was quantile normalized, the loadings between views are comparable.}
\label{sup_fig_loadings}
\end{figure}

\begin{figure}[h!]
\centering
\includegraphics[width=1\textwidth]{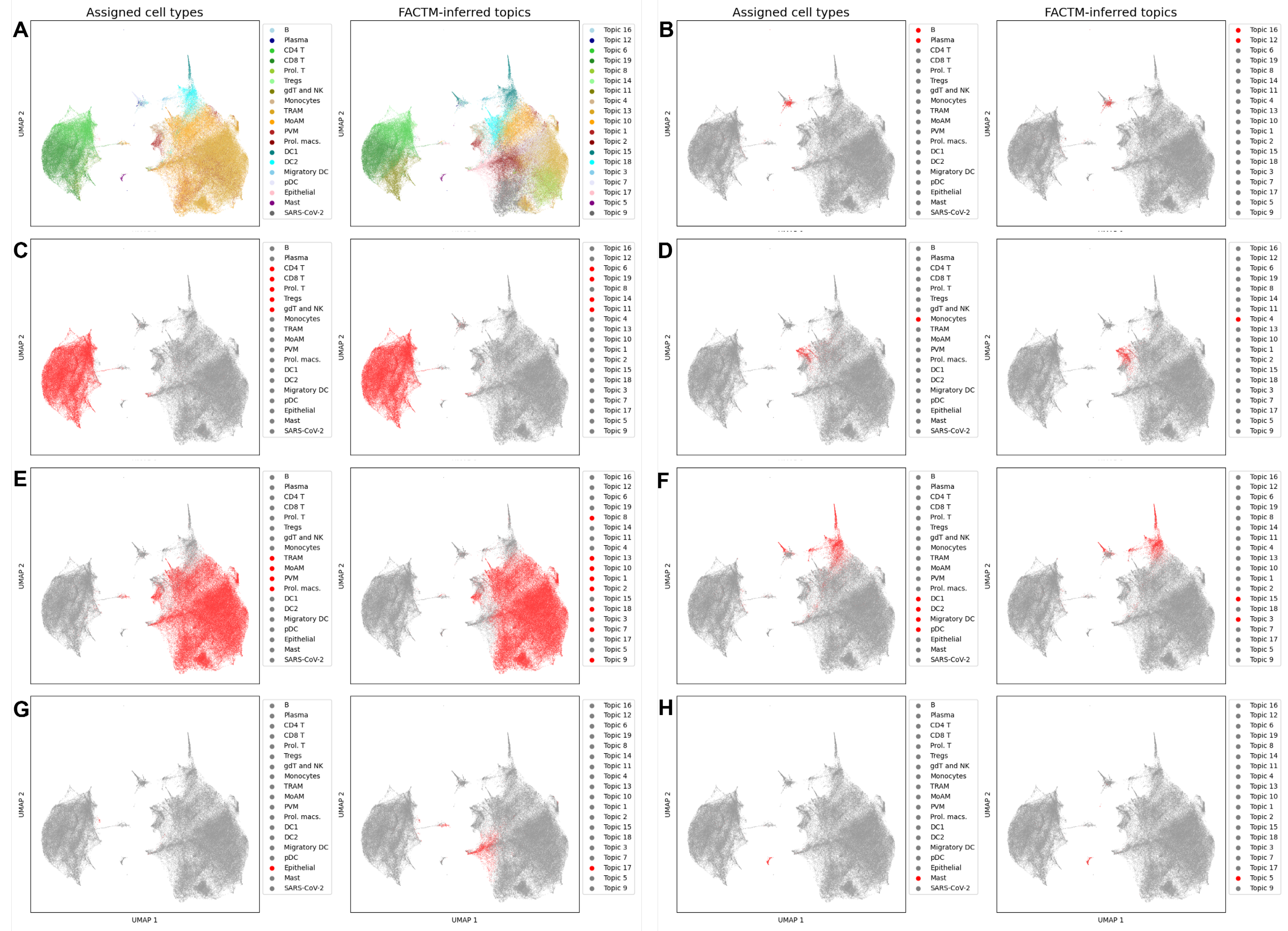}
\caption{UMAP plots comparing the clusters inferred by FACTM with the cell types assigned in the original source of the data. The UMAP representation is based on a matrix of 1,000 highly variable genes, with points colored according to the assigned clusters/cell types. Each point represents a single cell. \textbf{A.} Cell types and their corresponding topics, as in Fig. 7. \textbf{B. - H.} Comparison of more general cell types and the assigned clusters. Points are colored by selected cell types and topics (each topic is colored red once).}
\label{sup_fig_loadings}
\end{figure}

\clearpage
\section{ADDITIONAL TABLES}
\setcounter{table}{0}
\renewcommand{\thetable}{C.\arabic{table}}
\begin{table}[h!]
\caption{Supplement to Table 1 from the main text. Performance of the random forest classifier
using inferred latent factors
across the datasets for predicting assigned labels. Results are presented as the mean
ROC-AUC and PR-AUC ± standard deviation over 10-fold cross-validation.} \label{sup_tab_fa}
\centering
\begin{tabular}{cl|lll}
  \hline
Measure & Model & Mirex & CMU-Mosei & CMU-Mosi \\ 
  \hline
\multirow{4}{*}{PR-AUC} & FA+CTM & $0.39\pm 0.04$ & $0.70\pm 0.01$ & $0.66\pm 0.05$ \\ 
  & FACTM & $0.39\pm 0.03$ & $0.70\pm 0.01$ & $\textbf{0.67}\pm 0.04$ \\ 
  & MOFA & $0.39\pm 0.04$ & $\textbf{0.71}\pm 0.01$ & $0.65\pm 0.05$ \\ 
& muVI & $\textbf{0.40}\pm 0.03$ & $0.67\pm 0.01$ & $\textbf{0.67}\pm 0.04$ \\ 
  \hline
 \multirow{4}{*}{ROC-AUC} & FA+CTM & $\textbf{0.68}\pm 0.04$ & $\textbf{0.73}\pm 0.01$ & $0.67\pm 0.04$ \\ 
  & FACTM & $0.67\pm 0.02$ & $\textbf{0.73}\pm 0.01$ & $\textbf{0.68}\pm 0.04$ \\ 
 & MOFA & $\textbf{0.68}\pm 0.03$ & $\textbf{0.73}\pm 0.01$ & $0.66\pm 0.04$ \\ 
 & muVI & $\textbf{0.68}\pm 0.02$ & $0.70\pm 0.01$ & $0.67\pm 0.04$ \\ 
   \hline
\end{tabular}
\end{table}

\begin{table}[h!]
\caption{Supplement to Table 2 from the main text. Performance of the random forest classifier
using model-specific observation-level representation from the structured part of data
across the datasets. Results are presented as the mean
ROC-AUC and PR-AUC ± standard deviation over 10-fold cross-validation.} \label{sup_tab_ctm}
\centering
\begin{tabular}{cl|lll}
  \hline
Measure & Model & Mirex & CMU-Mosei & CMU-Mosi \\ 
  \hline
\multirow{3}{*}{PR-AUC} & CTM & $0.29\pm 0.03$ & $0.52\pm 0.02$ & $0.56\pm 0.05$ \\ 
   & FACTM & $\textbf{0.35}\pm 0.06$ & $0.60\pm 0.01$ & $\textbf{0.60}\pm 0.05$ \\ 
   & LDA & $0.33\pm 0.06$ & $\textbf{0.63}\pm 0.02$ & $0.56\pm 0.04$ \\ 
   \hline
 \multirow{3}{*}{ROC-AUC} & CTM & $0.57\pm 0.04$ & $0.55\pm 0.02$ & $0.57\pm 0.05$ \\ 
   & FACTM & $\textbf{0.64}\pm 0.06$ & $0.63\pm 0.01$ & $\textbf{0.61}\pm 0.06$ \\ 
  & LDA & $0.62\pm 0.05$ & $\textbf{0.66}\pm 0.01$ & $0.57\pm 0.04$ \\ 
   \hline
\end{tabular}
\end{table}

\section{NOTATION}
\renewcommand{\thetable}{D.\arabic{table}}

\paragraph{Distributions}
We use the following notation to represent  probability distributions:
\begin{itemize}
  \item \( \mathcal{N}(\mu, \sigma^2) \) - normal distribution (\(\mu\) - mean, \(\sigma^2\) - variance)
  \item \( \mathcal{N}_L(\mu, \Sigma) \) - multivariate normal distribution (\(\mu\) - mean vector, \(\Sigma\) - covariance matrix)
  \item \( \text{Bern}(p) \) - Bernoulli distribution (\(p\) - success probability)
  \item \( \mathcal{G}(a,b) \) - Gamma distribution (\(a\) - shape parameter, \(b\) - rate parameter)
  \item \( \text{Beta}(a,b) \) - Beta distribution (\(a\) - shape parameter, \(b\) - shape parameter)
  \item \( \text{Dir}(\alpha) \) - Dirichlet distribution (\(\alpha\) - concentration parameter vector)
  \item \( \text{Mult}(n, p) \) - Multinomial distribution (\(n\) - number of trials, \(p\) - probability vector)
\end{itemize}

\paragraph{FACTM notation} Table \ref{tab:notation_table} provides an overview of the notation used in this article to describe the model.

\begin{table}[h]
\caption{Notation used in FACTM: Indices, observed variables, and latent variables.}
    \label{tab:notation_table}
    \centering
    \renewcommand{\arraystretch}{1.5}
    \begin{tabular}{>{\centering\arraybackslash}p{1.8cm}|p{8.5cm}|p{5cm}}
    \textbf{Notation} & \textbf{Meaning} & \textbf{Notes} \\
        \hline
        $N$   & Number of samples (observations) & \\
        $M$   & Number of single views (modalities) & \\
        $D^m$   & Number of features in modality $m$ & Can vary across modalities \\
        $K$   & Number of latent factors & Hyperparameter\\
        $L$   & Number of clusters (topics) & Hyperparameter\\
        $I_n$   & Number of data points in sample $n$ & Can vary across samples \\
        $J_i$ & Number of observed objects in data point $i$  & Can vary across data points \\
        $G$ & Number of distinct object types (e.g. distinct words)  & \\
        \hline
        $y_{n,d}^m$  & Feature $d$ for sample $n$ in modality $m$ & Observed \\
        $\overline{y}_{n,i,j}$ & {Object $j$  in data point $i$ of sample $n$\newline
        $\overline{y}_{n,i,j} \in \{1,2,\ldots,G\}$}&  Observed \\
        $\overline{y}_{n,i,g}$ & Number of objects of type $g$  in data point $i$ of sample $n$ \newline
        $\overline{y}_{n,i,g} = \sum_{j=1}^{J_i} \mathbb{I}(\overline{y}_{n,i,j} = g)$&  Observed \\
        $\{\overline{y}_{n,i,g}\}_{g=1}^{G}$ & Object counts by type in data point $i$ of sample $n$ & Observed \\
        \hline
        $z_{n,k}$   & Latent factor $k$ for sample $n$  & Latent factors \\
        $w_{d,k}^m$   & Loading of factor $k$ for feature $d$ in modality $m$ \newline
        $w_{d,k}^m = \tilde{w}_{d,k}^m s_{d,k}^m$
        & Loadings for simple views 
        \\
        $\tilde{w}_{d,k}^m$ & Normal component of loading $w_{d,k}^m$ & Loadings - ARD prior \\
        $\alpha_k^m$ & Precision parameter for $\tilde{w}_{\cdot,k}^m$\newline Hyperparameters: $a_0^{\alpha}$, $b_0^{\alpha}$ & Loadings - ARD prior \\
        $s_{d,k}^m$ & Binary component of loading $w_{d,k}^m$  & Loadings - spike-and-slab prior \\
        $\theta_{k}^m$ & Probability of success for $s_{\cdot,k}^m$ \newline
        Hyperparameter: $a_0^{\theta}$, $b_0^{\theta}$&  Loadings - spike-and-slab prior \\
        $\overline{w}_{l,k}$ & Loading of factor $k$ for cluster (topic) $l$ & Loadings for structured view \\
        $\overline{\alpha}_k$ & Precision parameter for $\overline{w}_{\cdot,k}$ \newline Hyperparameters: $\overline{a}_0^{\alpha}$, $\overline{b}_0^{\alpha}$& Loadings - ARD prior\\
        $\mu_{n,l}$ & Modification of  population-level mean $\mu^{(0)}$ for sample~$n$ and cluster $l$ (sample-specific modification of cluster proportions) \newline Hyperparameter: $t$ &  \\
        $\mu^{(0)}$,  $\Sigma^{(0)}$ & Population-level mean and covariance & Dimensions: $L$ and $L \times L$ \\
        $\eta_{n}$ & After $\textrm{softmax}$ transformation, cluster distribution for sample $n$ & Dimension: $L$ \\
        $\xi_{n,i}$ & Cluster assignment for data point $i$ in sample $n$ & \\
        $\beta_{l}$ & Object type (word) distribution for cluster (topic) $l$ \newline
        Hyperparameter: $\alpha_0^{\beta}$& Dimension: $G$ \\
    \end{tabular}
    
\end{table}
\end{document}